\documentclass[lettersize,journal]{IEEEtran}
\usepackage{amsmath,amsfonts, amssymb}
\usepackage[linesnumbered,ruled,vlined]{algorithm2e}
\usepackage{array}
\usepackage[caption=false,font=normalsize,labelfont=sf,textfont=sf]{subfig}
\usepackage{textcomp}
\usepackage{stfloats}
\usepackage{url}
\usepackage{verbatim}
\usepackage{graphicx}
\usepackage{cite}
\usepackage{bm}
\usepackage{gensymb}
\usepackage{dsfont}
\usepackage{multirow}
\usepackage{hyperref}
\hypersetup{
  colorlinks = true,
  linkcolor = blue,
  citecolor = blue,
  urlcolor = black
}
\usepackage{siunitx}
\usepackage{booktabs}

\begin{document}

\title{Kinematics-Aware Multi-Policy Reinforcement Learning for Force-Capable Humanoid Loco-Manipulation}

\author{
  Kaiyan Xiao$^{1,\dagger}$, Zihan Xu$^{1, \dagger}$, Cheng Zhe$^{1}$, Chengju Liu$^{1, 2}$, Qijun Chen$^{1, 2}$~\IEEEmembership{}
\thanks{$^{\dagger}$ equal contribution}
\thanks{$^{1}$ the Robot and Artificial Intellifence Lab, College of Electronic and Information Engineering, Tongji University, Shanghai 201804, China}% <-this % stops a space
\thanks{$^{2}$ State Key Laboratory of Autonomous Intelligent Unmanned System(Tongji University), Shanghai, China}
\thanks{\{xiaokaiyan, xuzihan\}@tongji.edu.cn}
}

\markboth{Journal of \LaTeX\ Class Files,~Vol.~14, No.~8, August~2021}%
{Shell \MakeLowercase{\textit{et al.}}: A Sample Article Using IEEEtran.cls for IEEE Journals}

\IEEEpubid{0000--0000/00\$00.00~\copyright~2021 IEEE}

\maketitle

\begin{abstract}
Humanoid robots, with their human-like morphology, hold great potential for industrial applications. However, existing loco-manipulation methods primarily focus on dexterous manipulation, falling short of the combined requirements for dexterity and proactive force interaction in high-load industrial scenarios. To bridge this gap, we propose a reinforcement learning-based framework with a decoupled three-stage training pipeline, consisting of an upper-body policy, a lower-body policy, and a delta-command policy. To accelerate upper-body training, a heuristic reward function is designed. By implicitly embedding forward kinematics priors, it enables the policy to converge faster and achieve superior performance. For the lower body, a force-based curriculum learning strategy is developed, enabling the robot to actively exert and regulate interaction forces with the environment. 
% To ensure robust whole-body coordination, a delta-command policy is introduced to compensate for end-effector pose deviations in the world frame induced by lower-body motion. 
To ensure robust whole-body coordination, a delta-command policy is employed to counteract vertical end-effector displacements in the world frame resulting from lower-body motion.
Extensive simulation and real-world experiments on the Unitree G1 humanoid robot validate the proposed framework, showcasing its capability to accomplish high-payload tasks such as walking while carrying a 4 kg object and pushing or pulling a cart with a total load of 112.8 kg.
\end{abstract}


\begin{figure*}[!t]
    \centering
    \includegraphics[width=\linewidth]{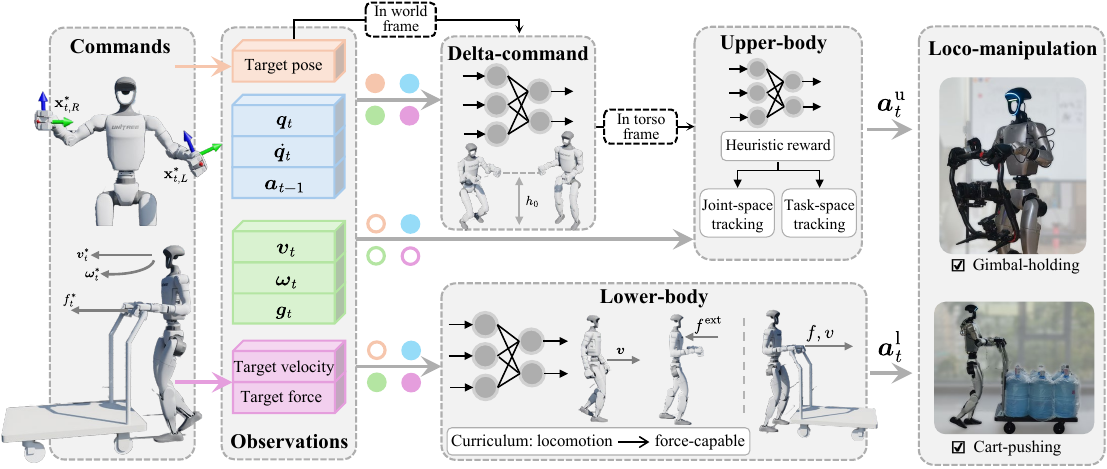}
    \caption{System architecture of the proposed training pipeline. The diagram illustrates the integration of the upper-body policy, lower-body policy, and delta-command policy for coordinated loco-manipulation tasks. When observation data are propagated, filled circles indicate the output of data in the corresponding color, whereas hollow circles indicate data not output. %$f^\text{ext}$ represents the external force exerted on the robot during lower-body training.
    }
    \label{fig: overview}
\end{figure*}

\section{Introduction}

\IEEEPARstart{H}{umanoid} robots are increasingly considered for deployment in industrial settings, where various tools and workflows are originally designed for human operators. As large-scale customization is often impractical, humanoid robots, owing to their morphology and natural operational compatibility, can seamlessly interface with and utilize existing tools. 

While humanoid robots have demonstrated considerable dexterous manipulation capabilities in low-load scenarios, such as tidying objects~\cite{fu2025humanplus}, delivery~\cite{ben2025homie}, and cleaning~\cite{kim2019control}, reliance on dexterity alone is insufficient when extending to labor-intensive industrial applications. In high-load automated production environments, common tasks include bimanual handling of heavy components~\cite{saeedvand2021hierarchical}, supporting or operating large-scale tools~\cite{liftting}, and transporting materials using carts~\cite{cart-pushing-icra}. These tasks typically require robots to possess both locomotion and manipulation capabilities, representing a canonical loco-manipulation problem.
However, unlike low-load manipulation tasks where precise motion dominates, high-load scenarios introduce significant dynamic coupling between the upper-body force exertion and lower-body stability, making whole-body coordination substantially more challenging.

% In recent years, reinforcement learning (RL) has demonstrated remarkable effectiveness in whole-body motion control for humanoid robots. 
% In~\cite{he2025attention-sr}, humanoids have achieved stable locomotion across various terrains, while motion capture (MoCap) data has been employed to impart highly expressive, human-like motion capabilities~\cite{he2025asap}.
% Stable locomotion across various terrains has been achieved in~\cite{he2025attention-sr}, while expressive, human-like motions have been synthesized using motion capture (MoCap) data in~\cite{he2025asap}.
Reinforcement learning (RL) has recently demonstrated strong potential for whole-body motion control in humanoid robots, enabling stable locomotion across diverse terrains~\cite{he2025attention-sr} as well as highly expressive, human-like behaviors through the use of motion capture (MoCap) data~\cite{he2025asap}. Whole-body RL-based motion control can be broadly categorized into two approaches: monolithic whole-body control~\cite{li2025amo, cheng2024expressive} and decoupled modular control~\cite{ben2025homie, zhang2025falcon,lu2024mobile}. Although monolithic whole-body control can theoretically generate natural and coordinated motions, it often struggles to simultaneously address both locomotion and manipulation tasks in high degree of freedom (DoF) humanoid robots, resulting in low training efficiency and insufficient exploration~\cite{ben2025homie}. Some works mitigate this issue by leveraging MoCap data to guide policy learning~\cite{he2024h2o,ji2025exbody2,zhang2025hub}. However, MoCap data inherently lack information about the force interactions between the robot and the environment, which limits the policy's ability to learn interaction-rich behaviors. 

By separating the robot into upper- and lower-body modules, decoupled controllers alleviate the challenges associated with high-dimensional optimization~\cite{zhang2025falcon}. Despite this advantage, several critical limitations remain. The upper-body typically relies on inverse kinematics for end-effector pose tracking~\cite{ben2025homie, lu2024mobile}, which becomes computationally intensive for high-DoF manipulators and struggles to meet real-time requirements~\cite{tian2021analyticalIK}. Its incompatibility with reinforcement learning frameworks further constrains the upper-body's generalization capability in industrial settings. Meanwhile, existing lower-body methods primarily focus on passive adaptation to external disturbances to ensure stable locomotion~\cite{zhang2025falcon, ben2025homie}. The ability to actively exert forces on the environment, which is a key requirement for high-load industrial tasks, has received little attention. Furthermore, achieving effective coordination between the upper- and lower-body modules within a decoupled framework remains largely unresolved~\cite{ding2025jaeger}. Collectively, these challenges highlight a critical gap in enabling humanoid robots to perform robust, high-load tasks in complex industrial scenarios.
\IEEEpubidadjcol

To address the aforementioned challenges, this work proposes a RL-based humanoid loco-manipulation framework, as illustrated in Fig. \ref{fig: overview}. The framework employs a three-stage training pipeline to endow humanoid robots with capabilities suitable for industrial applications. It operates within a command space defined by end-effector poses, locomotion velocity, and target exerted force, together with corresponding robot states as observations.
% In the framework, the command space includes the upper-body end-effector poses in the world coordinate frame, locomotion velocity, and target exerted force. These commands, combined with the robot's states, constitute the observation space. In the observation flow, filled circles indicate the inclusion of the corresponding observations, while hollow circles indicate their absence. 
% delta-command module
A key component of the framework is the delta-command module, which computes a compensatory offset for the end-effector pose in the torso frame. This offset mitigates vertical disturbances caused by torso motion during locomotion. The compensated pose serves as a reference for the upper-body policy, while the lower-body policy generates locomotion actions, together achieving coordinated whole-body control. 
% 启发式上肢奖励函数
Besides, during upper-body training, a heuristic reward function combining joint-angle errors and end-effector pose errors is designed. By implicitly embedding forward kinematics priors, the policy learns the mapping between joint configurations and end-effector poses rapidly. Moreover, in the lower-body training stage, we adopt a curriculum learning strategy to sequentially endow the policy with locomotion and force-capable skills. After the policy masters basic locomotion, target exerted forces are introduced, and external disturbances of equal magnitude but opposite direction to the target exerted force are applied to the robot in simulation. The policy is then trained to maintain stable locomotion under these conditions, gradually acquiring the ability to exert forces proactively.

% In the lower-body training stage, we introduce a force-based curriculum learning approach to develop force-capable locomotion. 
% In particular, external disturbances, equal in magnitude but opposite in direction to the target exerted force, are applied to the robot in simulation, and the policy is trained to maintain stable locomotion under such conditions. This gradually guides the policy to acquire the ability to exert forces proactively. 
% Notably, walking inevitably induces vertical body motion, which perturbs the upper-body end-effector poses. 
% To compensate for this effect, we introduce a delta-command policy to coordinate upper- and lower-body motions. 
% This policy operates in the world frame and reduces end-effector deviations from body motion by superimposing its commands on the upper-body policy
The main contributions of this work are summarized as follows:
\begin{enumerate}
  \item To enhance both training efficiency and performance of the upper-body, a heuristic reward function is designed, achieved by implicitly embedding forward kinematics priors.
  \item A force-based curriculum learning approach is proposed to endow the robot with force-capable skills, enhancing its adaptability in high-load and strong-interaction tasks.
  \item By introducing a delta-command policy, motion coordination under decoupled upper- and lower-body control is achieved.
  \item A three-stage RL framework for industrial loco-manipulation is proposed. By coordinating three-modular policies, the Unitree G1 robot achieves stable locomotion while carrying a \qty{4}{kg} object and pushing or pulling a cart with a load of \qty{112.8}{kg}.
\end{enumerate}

The remainder of this paper is organized as follows. Section~\ref{sec:related work} reviews the related literature. Section~\ref{sec:approach} introduces the proposed three-stage training pipeline. Section~\ref{sec:experiment} presents simulation and hardware experiments conducted to validate the effectiveness of the framework. Finally, Section~\ref{sec:conclusion} concludes the paper.

\section{Related Works}\label{sec:related work}

\subsection{Model-based Approaches for Loco-manipulation}
% Model-based loco-manipulation controllers have a long history of development, essentially formulating loco-manipulation control as an optimal control problem. Karim et al.~\cite{bouyarmane2018quadratic} formulate task-space multi-objective controllers as a unified quadratic program to model robots and their interactive entities as a centralized system, enabling physically consistent position and force control across humanoid manipulation tasks. The study~\cite{murooka2021humanoidroll} proposes a frequency-domain stabilization control method that explicitly accounts for external manipulation forces, using a pattern generator to track force references and adjusting center of mass and zero moment point to compensate for force tracking errors, demonstrated on tasks like object rolling. Li et al.~\cite{li2023dynamic} develops a model predictive control framework with a simplified rigid body dynamics model, demonstrating dynamic locomotion and loco-manipulation on the HECTOR humanoid, including load carrying and walking on uneven terrain. Although extensive progress has been made in improving the efficiency of optimal control solvers, their substantial online computational burden continues to be a major obstacle to their application in complex scenarios. Furthermore, related work~\cite{rioux2015humanoidNAO} decomposes locomotion and manipulation into action primitives on top of model-based walking, but this approach has yet to be validated on larger-scale robots. Another study~\cite{6386085} adjusts the contact locations based on measured forces during manipulation. 

Model-based loco-manipulation controllers have a long history, typically framing loco-manipulation as an optimal control problem~\cite{wensing2023optimization, 10375151, 10160523 }. Prior work has explored various strategies to enhance performance. Karim et al.\cite{bouyarmane2018quadratic} unify task-space multi-objective controllers into a quadratic program, enabling physically consistent position and force control across manipulation tasks. Murooka et al.\cite{murooka2021humanoidroll} propose a frequency-domain stabilization method that explicitly accounts for external forces, adjusting the center of mass and zero moment point to track force references. Li et al.\cite{li2023dynamic} develop a model predictive control framework based on a simplified rigid body model, demonstrating dynamic locomotion and manipulation on uneven terrain. Other works explore decomposition of locomotion and manipulation into action primitives\cite{rioux2015humanoidNAO} or adapt contact locations based on measured forces~\cite{6386085}.

% Model-base 方法的现实局限性
Despite these advances, applying model-based controllers in real-world scenarios remains challenging. They rely on accurate system dynamics and environment models, making them sensitive to modeling errors and unpredictable variations in contact forces, payloads, friction, and sensor noise. As a result, their robust performance is often compromised outside controlled settings, limiting practical deployment in complex industrial environments.

% However, in real-world scenarios, the contact force environment is highly variable, and sensor anomalies can severely compromise controller stability. Model-based approaches inherently rely on accurate system dynamics and environment models, making them sensitive to modeling inaccuracies and unstructured environment changes. In real-world industrial settings, where payloads, friction coefficients, and contact geometries often vary unpredictably. These controller may fail to maintain robust performance, limiting their practical deployment despite strong performance in controller environments.

\subsection{RL for Humanoid Locomotion}
Model-based methods are theoretically capable of generating stable gaits, yet their real-world application is hindered by poor adaptation to complex terrains and external disturbances. 
As a result, reinforcement learning (RL) has emerged as a more promising paradigm for tackling the challenges of humanoid locomotion control.
Lee~et~al.~\cite{lee2020leggedlocomotion} established the fundamental principles of generating stable gaits for legged robots through reinforcement learning.
% In~\cite{lee2020leggedlocomotion}, the fundamental principles of generating stable gaits for legged robots through RL were established. 
Building upon this foundation, subsequent works introduced domain randomization to enable sim-to-real transfer of RL policies~\cite{tan2018sim}, achieving robust performance on physical humanoid robots~\cite{radosavovic2024real}. 
Meanwhile, the emergence of large-scale parallel simulation platforms such as IsaacSim~\cite{mittal2023orbit} and MuJoCo~\cite{zakka2025mujoco} has made large-scale RL training feasible, further advancing RL-based progress and breakthroughs in robotic locomotion control. In recent years, RL-powered humanoid robots have not only maintained stable locomotion under diverse terrains and external perturbations, but have also demonstrated the capability to perform more complex skills, such as jumping~\cite{li2023jumping}, stair climbing~\cite{sun2025stairclimbing}, and even parkour~\cite{zhuang2024parkour}.

% RL 行走技术完善，构建了很好的基础
These advances collectively indicate that RL-based humanoid locomotion control has matured significantly, demonstrating robustness, reproducibility, and real-world deployability across diverse conditions. This maturity establishes a solid foundation upon which integrated loco-manipulation frameworks, proposed in this paper, can effectively built.

\subsection{RL-based Whole-Body Control for Loco-manipulation}
With the advancement of RL-based locomotion methods, RL has also been applied to the training of whole-body controllers for humanoid~\cite{dao2024box, li2021reinforcement, 10375203, zhang2025wococo}. These controllers are typically categorized into monolithic control and decoupled modular control. Monolithic control often relies on motion-capture data as reference trajectories. AMP~\cite{peng2021amp} incorporates expert priors and employs adversarial learning to generate natural motions. H2O~\cite{he2024h2o}, on the other hand, leverages large-scale motion-capture datasets to enable dexterous manipulation. Ze et al.~\cite{ze2025twist} present a teleoperation system that combines RL and behavior cloning to achieve real-time, coordinated whole-body control. Nevertheless, such approaches are limited to motion imitation and fail to learn active force interactions with the environment. 

In decoupled control frameworks, Ben et al.~\cite{ben2025homie} employ inverse kinematics and PD controllers for upper-body control, while utilizing a RL policy for lower-body locomotion, and further design an isomorphic exoskeleton to facilitate intuitive teleoperation by human operators. Lu et al.~\cite{lu2024mobile} adopt a similar modular architecture and introduce PMP, a learned latent representation of upper-body motion, to condition the lower-body RL policy, thereby ensuring dynamic stability and coordination in whole-body tasks. While these approaches leverage inverse kinematics to achieve dexterous upper-body manipulation, their reliance on kinematic control inherently limits force generation and adaptability, rendering them unsuitable for high-load industrial scenarios. The study~\cite{zhang2025falcon} further incorporates force interactions through a force curriculum that jointly trains upper- and lower-body policies, equipping the robot with force-adaptive capabilities. Nevertheless, the interaction remains reactive rather than proactive, limiting the robot's ability to actively exert and regulate forces in dynamic environments.

\subsection{Our Approach}
% The proposed loco-manipulation framework is inspired by several prior works. For upper-body training, the study~\cite{xu2025NAOnpr} employs a joint angle signal network to map joint angles to end-effector poses. We extend this approach from an offline to an online formulation and draw inspiration from the bilevel optimization method in the research~\cite{xie2025kungfubot} for tuning the parameters of the heuristic reward functions during training. For lower-body training, we build upon the training pipeline of the study~\cite{sun2025stairclimbing} and extend the approach in the prior work~\cite{zhang2025falcon}, which achieves force-adaptive capability by applying forces at the robot's end-effector. During training, force commands are introduced and reaction forces corresponding to the desired output are applied to the robot's body, endowing the robot with the ability to actively exert forces on the environment during real-world deployment.
% Finally, while He et al.~\cite{he2025asap} introduces a delta-action module to bridge the gap between simulation and reality, we propose a delta-command module to address the mismatch between task-space and operational space.

The proposed loco-manipulation framework draws inspiration from several previous studies. For upper-body training, the study by Xu et al.~\cite{xu2025NAOnpr} employs a joint angle signal network to map joint angles to end-effector poses. We extend this approach from offline to online training and further adopt the bilevel optimization strategy from Xie et al.~\cite{xie2025kungfubot} to tune the parameters of heuristic reward functions during training. For lower-body training, we build upon the training pipeline of the study~\cite{sun2025stairclimbing} and extend the force-adaptive approach presented by Zhang et al.\cite{zhang2025falcon}, which applies forces at the robot's end-effectors to enable adaptive interaction. During our training pipeline, force commands are introduced and reaction forces corresponding to the target exerted force are applied to the robot's torso, equipping the robot with the ability to actively exert forces on the environment during real-world deployment. Finally, while the work of He et al.~\cite{he2025asap} introduces a delta-action module to reduce the simulation-to-reality gap, we propose a delta-command module to address discrepancies between task-space and operational-space commands.

\section{Approach}\label{sec:approach}
% 人形机器人 loco-manipulation 任务需要上肢和下肢的协同操作。

As illustrated in Fig.~\ref{fig: overview}, the proposed framework consists of three policies. Section~\ref{sec:problem formulation} defines the problem, Section~\ref{sec:upper-body} presents the upper-body policy, Section~\ref{sec:lower-body} describes the lower-body policy, and Section~\ref{sec:delta-command} introduces the delta-command policy.

\subsection{Problem Formulation}\label{sec:problem formulation}
The Unitree G1 humanoid robot is adopted as the experimental platform, which features a total of $n=29$
actuated joints. The DoFs are partitioned into $n^{\text{l}}=15$ lower-body joints (waist, hip, knee, ankle) and $n^{\text{u}}=14$ upper-body joints (shoulder, elbow, wrist).

We formulate the humanoid loco-manipulation problem as a goal-conditioned RL task. The problem is defined by a Markov Decision Process (MDP):
\begin{equation}
\mathcal{M}=(\mathcal{S}, \mathcal{A}, \mathcal{G}, \mathcal{P}, \mathcal{R}, \gamma),
\end{equation}
where $\mathcal{S}$ is the state space of the humanoid robot, $\mathcal{A}$ is the robot's action space, $\mathcal{G}$ is the goal space, $\mathcal{P}$ is the state transition probability function, $\mathcal{R}$ is the reward function and $\gamma \in [0,1)$ is the discount factor. 

At each time step $t$, upper-body proprioception $s^{\text{u}}_{t}$ and lower-body proprioception $s^{\text{l}}_{t}$ are aggregated into the robot's overall proprioceptive state $s_{t} \in \mathcal{S}$.
\begin{equation}
\begin{aligned}
  s^{\text{u}}_{t}&=[\bm{q}^{\text{u}}_{t}, \bm{\dot{q}}^{\text{u}}_{t}, \bm{a}^{\text{u}}_{t-1}], \\
  s^{\text{l}}_{t}&=[\bm{q}^{\text{l}}_{t}, \bm{\dot{q}}^{\text{l}}_{t}, \bm{\omega}_{t}, \bm{g}_{t}, \bm{a}^{\text{l}}_{t-1}], \\
  s_{t}&=[\bm{q}_{t}, \bm{\dot{q}}_{t}, \bm{\omega}_{t}, \bm{g}_{t}, \bm{a}_{t-1}],
\end{aligned}
\end{equation}
where $\bm{q}_{t} \in \mathbb{R}^{n}$ and $\bm{\dot{q}}_{t} \in \mathbb{R}^{n}$ are the joint positions and velocities, $\bm{\omega}_{t} \in \mathbb{R}^{3}$ is the root angular velocity, $\bm{g}_{t} \in \mathbb{R}^{3}$ is the projected gravity, and $\bm{a}_{t-1} \in \mathbb{R}^{n}$ is the action in the last time step.

The goal space $\mathcal{G}_{t}$ consists of manipulation goals $\mathcal{G}^{\text{u}}_{t}$, locomotion goals $\mathcal{G}^{\text{l}}_{t}$, and high-level coordination goals $\mathcal{G}^{\text{h}}_{t}$.
\begin{equation}
\begin{aligned}
  \mathcal{G}^{\text{u}}_{t}&=\{\mathbf{x}^{\text{torso}*}_{t, i}\}_{ i \in  \{\text{L}, \text{R}\}},\\
  \mathcal{G}^{\text{l}}_{t}&=[\bm{v}^{*}_{t}, \bm{\omega}^{*}_{t}, f^{*}_{t}], \\
  \mathcal{G}^{\text{h}}_{t}&=\{\mathbf{x}^{\text{world}*}_{t, i}\}_{ i \in  \{\text{L}, \text{R}\}}.
\end{aligned}
\end{equation}
Let the subscripts ``$\text{L}$'' and ``$\text{R}$'' denotes the left and right end-effectors of the upper-body, $\mathbf{x}^{j}$ is defined as the end-effector's pose relative to the $j$-frame, represented in Cartesian coordinates and Euler angles. $\bm{v}_{t}$, $\bm{\omega}_{t}$ and $f_{t}$ represent root linear velocity, root angular velocity, and the magnitude of forces applied to the environment, respectively. Throughout this paper, the superscript ``$*$'' is used to indicate the target (or commanded) state of the robot.

\begin{algorithm}[!t]
\caption{Sampling End-effector Poses}
\label{alg:sampling joint pose}
\DontPrintSemicolon
\KwIn{Left end-effector's task-space $\mathcal{W}_{\text{L}}$, left arm joints' upper and lower limits $\bm{q}^{\text{u},\max}_{\text{L}}$, $\bm{q}^{\text{u},\min}_{\text{L}}$}
\KwOut{$\bm{q}^{\text{u}*}$, $\mathbf{x}^{\text{torso}*}_{\text{L}}$, $\mathbf{x}^{\text{torso}*}_{\text{R}}$}

\Repeat{$\mathbf{x}^{\textup{torso}*}_{\textup{L}} \in \mathcal{W}_{\textup{L}}$ \textbf{and} $\vartheta < \frac{\pi}{2}$}{
  $\bm{u} \gets  (u_{0},...,u_{\frac{n^{\text{u}}}{2}}), \quad u_{i} \sim \mathcal{U}(0,1)$\;
  
  $\bm{q}^{\text{u}*}_{\text{L}} \gets \bm{q}^{\text{u}, \min}_{\text{L}} + \bm{u} \odot (\bm{q}^{\text{u},\max}_{\text{L}} - \bm{q}^{\text{u}, \min}_{\text{L}})$\;
  
  %Forward kinematics: \\
  $\mathbf{x}^{\text{torso}*}_{\text{L}} \gets \mathrm{FK}(\bm{q}^{\text{u}*}_{\text{L}})$ \tcp*{\small $\text{FK}(\cdot)$ represents forward kinematic computation.}
  
  Compute left end-effector's direction vector $\bm{e}_{\text{L}}$\;
  
  Compute left elbow link's direction vector $\bm{e}_{\text{elbow}}$\;
  
  % Compute angle: \\
  $\vartheta \gets \cos^{-1}\!\left(\dfrac{\bm{e}_{\text{L}} \cdot \bm{e}_{\text{elbow}}}{\|\bm{e}_{\text{L}}\| \|\bm{e}_{\text{elbow}}\|}\right)$\;
}
% Mirror left joint angles to right arm: \\
$\bm{q}^{\text{u}*}_{\text{R}} \gets \mathrm{Mir}(\bm{q}^{\text{u}*}_{\text{L}})$ \tcp*{\small $\text{Mir}(\cdot)$ mirrors left joint angles to right arm.}

% Compute right arm forward kinematics: \\
$\mathbf{x}^{\text{torso}*}_{\text{R}} \gets \mathrm{FK}(\bm{q}^{\text{u}*}_{\text{R}})$\;

% Concatenate full configuration: \\
$\bm{q}^{\text{u}*} \leftarrow [\bm{q}^{\text{u}*}_{\text{L}}, \bm{q}^{\text{u}*}_{\text{R}}]$\;

\Return{$\bm{q}^{\textup{u}*}$, $\mathbf{x}^{\textup{torso}*}_{\textup{L}}$, $\mathbf{x}^{\textup{torso}*}_{\textup{R}}$}
\end{algorithm}

In the framework depicted in Fig.~\ref{fig: overview}, the goal space (commands) include the upper-body end-effector poses in the world coordinate frame, locomotion velocity, and target exerted force. Together with the robot's states, these commands constitute the observation space, where filled circles indicate the inclusion of the corresponding observations and hollow circles indicate their absence. Based on these definitions, our approach decouples the whole-body control problem into three separate policies. The high-level delta-command policy $\pi^{\text{h}}$ does not directly participate in robot control. The basic upper-body agent learns a policy $\pi^{\text{u}}$: $s^{\text{u}}_{t} \times \mathcal{G}^{\text{u}}_{t} \mapsto \bm{a}^{\text{u}}_{t}$, while the lower-body agent learns a policy $\pi^{\text{l}}: s^{\text{l}}_{t} \times \mathcal{G}^{\text{l}}_{t} \mapsto \bm{a}^{\text{l}}_{t}$. Each policy is optimized to maximize the expected cumulative reward:
\begin{equation}
  \max_{\theta_{i}}\mathbb{E}[\sum_{t=1}^{T}\gamma_{t-1} \cdot r^{i}_{t}], \quad
  r^{i}_{t} = \mathcal{R}^{i}(s^{i}_{t}, \mathcal{G}^{i}_{t}), \quad i \in \{\text{u}, \text{l}, \text{h}\},
\end{equation}
where $\theta_{i}$ denotes the parameters of the policy $i$, and $T$ is the episode length. 
$r^{i}_{t}$ is the scalar reward at timestep $t$, computed from the agent's current state $s^{i}_{t}$ and its task goal $\mathcal{G}^{i}_{t}$ through the reward function $\mathcal{R}^{i}(\cdot)$. 
$\gamma$ is the discount factor that balances immediate and future rewards.
% The control problem is decoupled into three reinforcement learning sub-policies. A high-level delta-command policy $\pi_{\Delta}: \left \langle s^{p}_{t}, \bm{a}^{\Delta}_{t-1} \right \rangle  \times \mathcal{G}^{\Delta}_{t} \Rightarrow \bm{a}^{\Delta}_{t} \in \mathbb{R}^{12}$ refines upper-body commands for robust whole-body coordination. The basic upper-body policy $\pi_{u}:  s^{p, u}_{t} \times \left \langle \mathcal{G}^{u}_{t}, \bm{a}^{\Delta}_{t} \right \rangle \Rightarrow \bm{a}^{u}_{t}$  and lower-body policy $\pi_{l}: s^{p, l}_{t} \times \mathcal{G}^{l}_{t} \Rightarrow \bm{a}^{l}_{t}$ generate upper-body actions and lower-body actions, respectively. 

During execution, the combined action $\bm{a}_{t}$ from upper- and lower-body policies is directly mapped to the desired joint positions $\bm{q}_{t}^{\text{des}}$:
\begin{equation}
  \begin{aligned}
  \bm{a}_{t} &= [\lambda^{\text{u}} \bm{a}^{\text{u}}_{t},
                    \lambda^{\text{l}} \bm{a}^{\text{l}}_{t}], \\
  \bm{q}_{t}^{\text{des}} &= 
                    \bm{a}_{t} + \bm{q}^{\text{default}}.
  \end{aligned}
\end{equation}
Where $\lambda^{\text{u}}$ and $\lambda^{\text{l}}$ are the action scaling factor for upper- and lower-body, $\bm{q}^{\text{default}}$ represents the default joint positions. $\bm{q}_{t}^{\text{des}}$  is then tracked by a joint-level PD controller, with the resulting motor torques $\bm{\tau}_{t}$ computed as:
\begin{equation}
  \bm{\tau}_{t} = \bm{K}_{p}(\bm{q}_{t}^{\text{des}} - \bm{q}_{t}) - \bm{K}_{d} \cdot \dot{\bm{q}_{t}},
\end{equation}
where $\bm{K}_{p}$ and $\bm{K}_{p}$ are diagonal gain matrices of stiffness and damping.

\subsection{Heuristic Upper-body Policy Optimization} \label{sec:upper-body}

\begin{figure}[!t]
\centering
\subfloat[]{\includegraphics[width=0.32\linewidth]{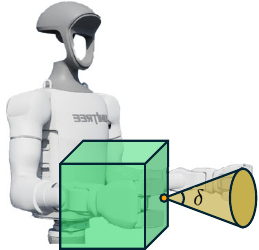}\label{fig: task-space}}
\qquad \quad
\subfloat[]{\includegraphics[width=0.32\linewidth]{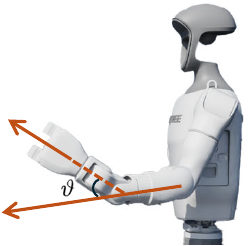}\label{fig: elbow filter}}
\caption{Task-Space and Imposed Constraints of the Upper Body. (a) Task-space of the upper body. The Task-spaces of the left and right end-effectors are symmetric with respect to the $x$-axis of the torso frame. The green cuboid illustrates the set of feasible positions, whereas the yellow cone indicates the feasible orientation range of the end-effector, with the aperture angle $\delta$ defining the orientation limits. (b) Illustration of the angle between the elbow link's direction and the end-effector's direction.}
\label{fig: task-space and elbow filter}
\end{figure}

% TODO: 上肢的精准控制是人形机器人进行loco-manipulation任务的前提。G1机器人有两条7轴机械臂，有着较大的工作空间。针对在工作空间进行位姿采样的强化学习方法而言，连续且庞大的工作空间会导致数据收集效率低下，进而使得策略的收敛速度慢。对此我们提出一种启发式上肢训练方法，有效加速策略训练过程，并提升末端追踪精读。

% 引入上身控制，说明工作空间采样策略收敛慢。
% 使用 heuristic reward来加速策略收敛，heuristic reward具体是什么。如何采样，mirror
% 结合 angle track 和 pose track，并说明std参数自适应

Precise control of the upper-body is a prerequisite for humanoid robots to perform loco-manipulation tasks. The Unitree G1 robot is equipped with two 7-DoF arms, providing a large workspace. Direct applying RL over such a continuous and high-dimensional workspace often results in inefficient data collection and slow policy convergence. To address this issue, a heuristic method that leverages kinematic priors to guide policy convergence is designed.

% 首先如下式，采样一组关节数据。
% \begin{equation}\label{joint sample}
%   \begin{aligned}
%     \bm{u} = (u_{0},...,u_{n^{u}}), u_{i} \sim \mathcal{U}(0, 1), i=0, ..., n^{u} \\
%     \bm{q}^{u}_{target} = \bm{q}^{u}_{\min} + \bm{u} \odot (\bm{q}^{u}_{\max} - \bm{q}^{u}_{\min})
%   \end{aligned}
% \end{equation}
% where $\mathcal{U}(a,b)$ denotes the uniform distribution over the interval [a,b], the operator $\odot$ denotes element-wise multiplication, ${q}^{u}_{\max}$, ${q}^{u}_{\min}$ are the vectors of the maximum and minimum joint positions of the upper-body. Subsequently, the end-effector Cartesian position $p_{\text{target}}$ and quaternion orientation $q_{target}$ in torso frame are obtained through forward kinematics based on the sampled joint configuration. %采样过程中，我们对结果进行了筛选，首先是末端位姿要落在设定的任务空间中，齐次elbow link的方向与末端方向的夹角要小于90°，这样可以有效提升数据的有效性（帮我想几个有点）。在此基础上，启发式奖励函数设计入下：

The training pipeline of the upper-body policy is presented as follows. End-effector poses are sampled using Alg.~\ref{alg:sampling joint pose}. In the sampling process, the poses are filtered to improve data quality and prevent non-physical postures. Specifically, the end-effector position must lie within the predefined task-space (as shown in Fig.~\ref{fig: task-space and elbow filter}\ref{sub@fig: task-space}), and the angle between the homogeneous elbow link direction and the end-effector direction must be less than \SI{90}{\degree} (as shown in Fig.~\ref{fig: task-space and elbow filter}\ref{sub@fig: elbow filter}). Furthermore, a mirror trick is applied to prevent performance inconsistency between the left and right hands caused by random sampling.

Based on the sampling method, the heuristic reward function $R^{\text{u}}$, which implicitly encodes kinematic information, is formulated as: 
\begin{equation}\label{eq:heuristic reward}
    R^{\text{u}} = \alpha^{\text{joint}} \cdot r^{\text{joint}} + \alpha^{\text{pose}} \cdot r^{\text{pose}}.
\end{equation}
To simultaneously track the objective pose in both the joint-space and the task-space, $R^{\text{u}}$ is decomposed into two terms, where $\alpha^{\text{joint}}$ and $\alpha^{\text{pose}}$ are weighting factors balancing their respective contributions. The joint-space term measures the discrepancy between the current joint position and the target joint positions, as expressed below:
\begin{equation}\label{eq:joint position term}
  r^{\text{joint}} = \exp(-\frac{\left \| \bm{q}^{u} - \bm{q}^{u*}\right \|}{\sigma^{2}}),
\end{equation}
 where $\sigma$ is a scaling parameter that controls the sensitivity of the joint-space exponential term. In contrast, the task-space term is formulated to directly capture the difference between the current end-effector pose and the target pose, as shown in the following equation:
\begin{equation}
  \begin{aligned}
    \Delta \phi &= \sum_{i \in \{\text{L},\text{R}\}} || \text{Log}(\eta^{\text{torso}}_{i} \otimes \hat{\eta}^{\text{\text{torso}*}}_{i}) ||, \\
     r^{\text{pose}} &= \left \| \mathbf{d}^{\text{torso}*}_{i} - \mathbf{d}^{\text{torso}}_{i} \right \|_{i \in \{\text{L},\text{R}\}} + \frac{1}{2} \Delta \phi,
  \end{aligned}
\end{equation}
Where $\eta^{j}$ denote the orientation of the end-effector in $j$-frame  as quaternion form, $\hat{\eta}$ is the conjugate of the quaternion, the operator $\otimes$ denotes the quaternion multiplication, $\text{Log}(\cdot)$ converts a quaternion into axis-angle representation, $\mathbf{d}^{k}$ represents the end-effector's position in $k$-frame, and $\Delta \phi$ is the orientation error. 

%训练初期 rjoint 的权重会更大，让运动学启发信息引导agent快速追踪至目标位置附近，但是由于关节角度误的差累计会导致末端追踪效果精确。因此，当策略收敛有收敛趋势时将rpose项的权重调大，让agent脱离运动学信息的指引，去追寻空间位姿。

In the context of the heuristic reward function, the weighting factors $\alpha^{\text{joint}}$ and $\alpha^{\text{pose}}$, as well as the scale parameter $\sigma$, are dynamically adjusted to guide the learning process. In the early stage, a higher weight is assigned to the joint-space term $r^{\text{joint}}$, allowing the forward kinematic information embedded in the reward function to guide the agent efficiently toward the vicinity of the target poses. As $\sigma$ gradually decreases, the robot's joint tracking error is further reduced, which facilitates rapid convergence by reducing inefficient exploration in the sparse-reward task-space. As the policy begins to converge, the weight of the task-space term $r^{\text{pose}}$ is increased, encouraging the agent to refine its end-effector trajectories and achieve precise task-space tracking. 

%Moreover, the exponential formulation of the joint-position tracking term prevents excessively large penalties for significant deviations, ensuring training stability while preserving sufficient exploration during the initial learning phase.

\subsection{Force-capable Lower Body Curriculum Learning}
\label{sec:lower-body}
\begin{figure}[!t]
\centering
\subfloat[]{\includegraphics[width=0.47\linewidth]{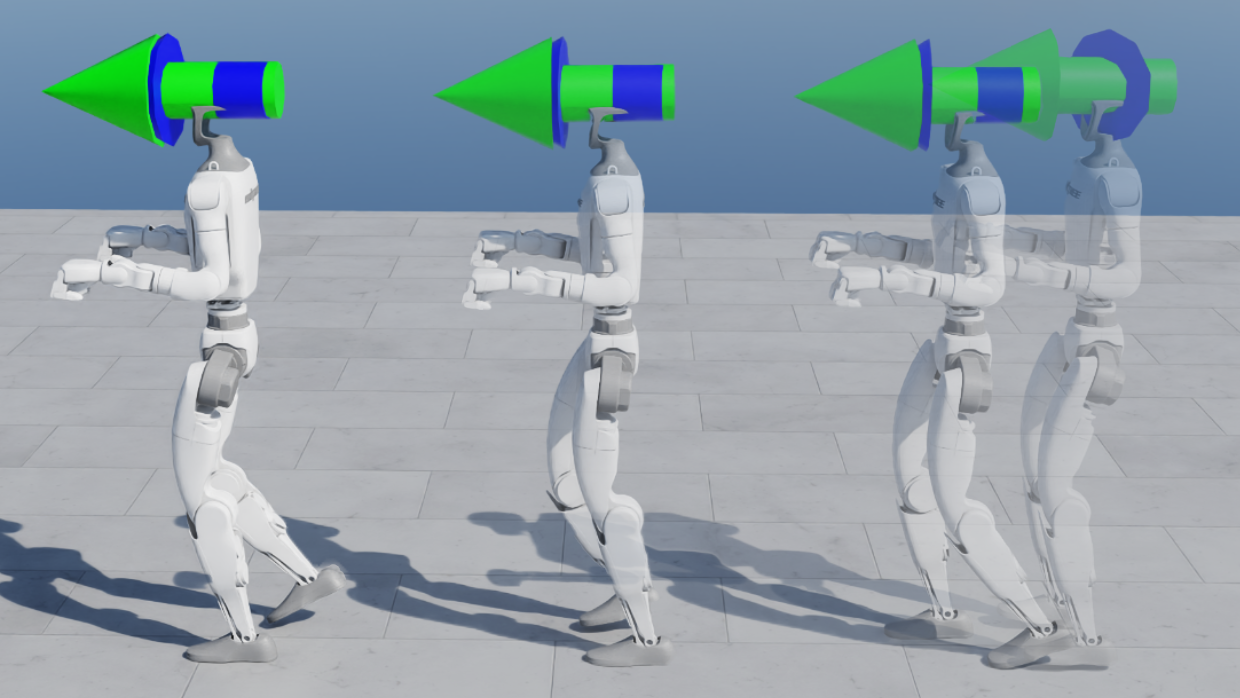}\label{fig: with_sample_method}}
\hfill
\subfloat[]{\includegraphics[width=0.47\linewidth]{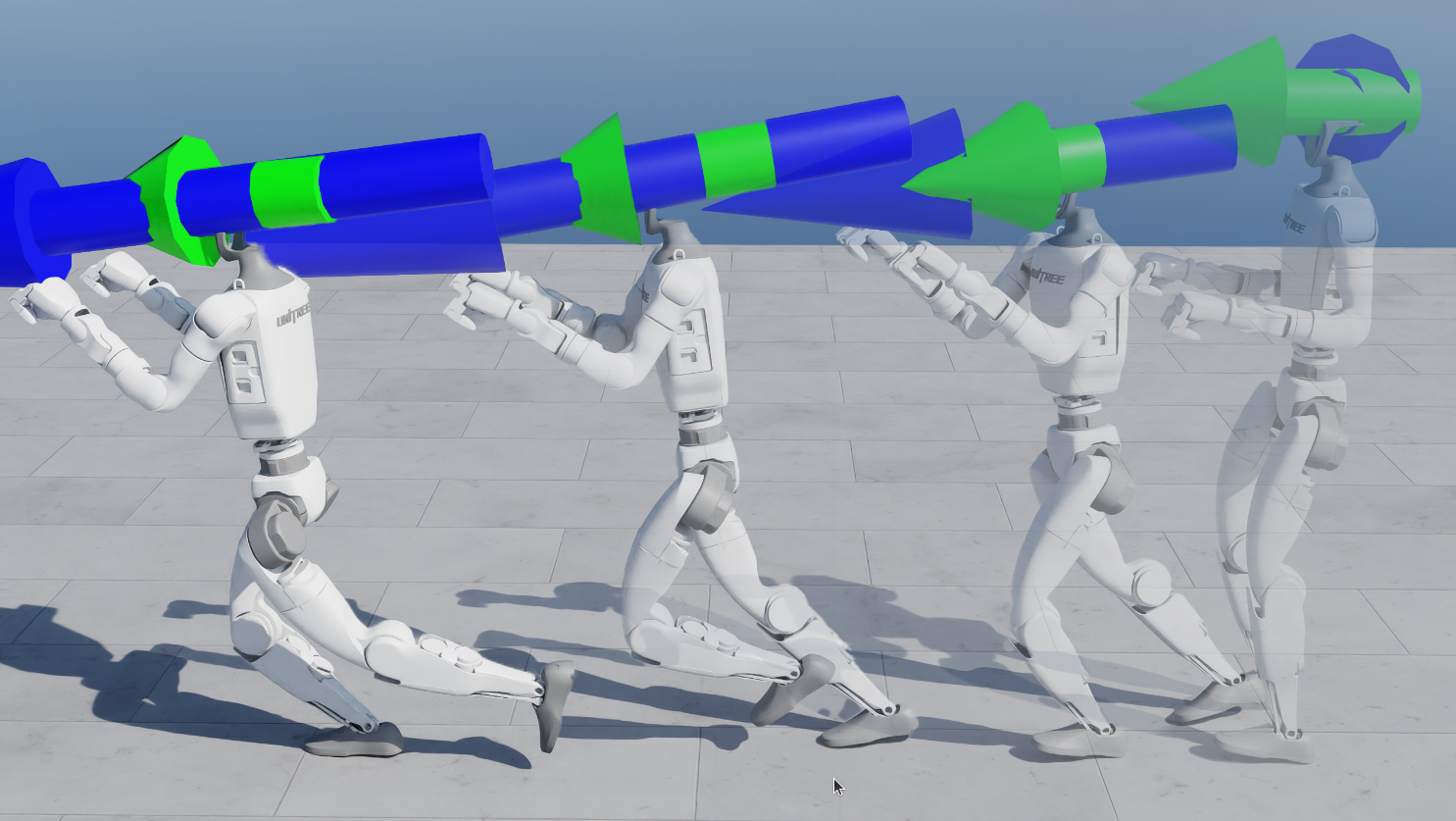}\label{fig:without_sample_method}}
\caption{Comparison of training outcomes with and without the sampling strategy. Both (a) and (b) are trained using curriculum learning, but only (a) incorporates the sampling strategy. Green arrows denote target velocities, and blue arrows denote actual velocities. The target linear velocity is set to $v^{*}_{t} = 0.7$~\si{\meter\per\second}, while the target force is $f^{*}_{t} = 0$~\si{\newton}. Without the sampling strategy, the policy tends to converge toward high-force scenarios, resulting in degraded locomotion performance under low-force conditions. In contrast, the agent trained with the sampling strategy exhibits improved generalization to varying force levels.}
\label{fig: compare of force sample method}
\end{figure}
%下肢控制是主要负责机器人的移动能力，引入force-aware的力量训练可以使机器人获得对外界主动施加力的能力，有效提升机器人与外界的交互能力。
% 引入下肢运动控制，并说明我们的方法可以让机器人学会主动施加力
% 下肢训练是用课程学习来的，课程学习判断力量上涨的评判指标：1. dis_meet_requirement & lin_meet_requirement & ang_meet_requirement，后两者是很有必要的，不然可能会出现机器人被力推着移动一定距离导致满足条件
% 同时说明力需要进行随机采样（更新施加力量大小，比例为：0.1不施加力，0.4在[0, force]之间随机采样，0.5施加力大小为force），否则如果力量长时间不增长会导致策略收敛到局部最优解（适应大力气），附上课程学习伪代码。

Lower-body policy primarily governs the robot's locomotion capability. The proposed lower-body training pipeline diverges from existing methods by integrating force-capable training, which enables the robot to actively apply forces to the environment and thereby significantly enhancing its interaction capabilities.

To achieve the force-capable training objective, an additional force target $f^{*}_{t}$ is introduced in the lower-body goal space $\mathcal{G}^{\text{l}}_{t}$. The direction of $f^{*}_{t}$ is set opposite to the robot's linear velocity, while its magnitude is sampled from a dynamic range according to a designed procedure. Once $f^{*}_{t}$ is determined, the simulation environment applies an equal and opposite force to the robot's torso in accordance with Newton's third law, which is necessary to make the training effective. However, continuously applying external forces during locomotion training presents a significant challenge. If the target force $f^{*}_{t}$ exceeds the agent's locomotion capability, it can hinder or slow down convergence. To address this, curriculum learning is incorporated into the lower-body training, enabling the agent to sequentially acquire locomotion and force-capable skills. 

During the training of the lower-body policy, the force magnitude $f^{*}_t$ is sampled according to:
\begin{equation}\label{eq:force_sample_strategy}
f^{*}_{t} = 
\begin{cases}
    0, & \text{if } u \leqslant 0.1, \\
    \mathcal{U}\left(0, f^{\max}_{t}\right), & \text{if } 0.1 < u \leqslant 0.5, \\
    f^{\max}_{t}, & \text{if } u > 0.5,
\end{cases}
 u \sim \mathcal{U}(0,1)
\end{equation}
where $f^{\max}_t$ denotes the upper bound of the force at the current time step. Furthermore, the progression of the force curriculum is regulated by multiple evaluation metrics, which are summarized as follows:
\begin{equation}\label{eq: force update condition}
  \mathcal{C}=\mathcal{C}_{\text{lin}} \wedge \mathcal{C}_{\text{ang}} \wedge \mathcal{C}_{\text{dis}} \wedge \mathcal{C}_{\text{force}}
\end{equation}
where each metric is defined as follows:
\begin{equation}
  \begin{aligned}
    \mathcal{C}_{\text{lin}} &= \frac{1}{N_{\text{steps}}} \sum_{i=1}^{N_{\text{steps}}} \left \| \bm{v}^{*}_{i}-\bm{v}_{i} \right \|  <  \varepsilon_{\text{lin}}, \\
    \mathcal{C}_{\text{ang}} &= \frac{1}{N_{\text{steps}}} \sum_{i=1}^{N_{\text{steps}}} \left \| \bm{\omega}^{*}_{i}-\bm{\omega}_{i} \right \| <  \varepsilon_{\text{ang}}, \\
    \mathcal{C}_{\text{dis}} &= \left \| p^{xy}_{o}-p^{xy} \right \| >  \varepsilon_{\text{dis}}, \\
    \mathcal{C}_{\text{force}} &= 
    \begin{cases}
        1, & \text{if } f^{*}_{t}=f^{\max}_{t}, \\
        0, & \text{otherwise}.
    \end{cases}
  \end{aligned}
\end{equation}
% TODO 介绍C
Here, $N_{\text{steps}}$ indicates the maximum number of time steps within a command cycle. $p^{xy}_{o}$ and $p^{xy}$ represent the robot's world-frame positions at the beginning and the end of the command cycle, respectively. $\varepsilon_{\text{lin}}$, $\varepsilon_{\text{ang}}$ and $\varepsilon_{\text{dis}}$ are empirically defined threshold values. Among these metrics, $\mathcal{C}_{\text{lin}}$ and $\mathcal{C}_{\text{ang}}$ measure the velocity tracking accuracy during a command cycle, 
$\mathcal{C}_{\text{dis}}$ captures the traveled distance, and $\mathcal{C}_{\text{force}}$ reflects whether the sampled force command attains its current upper bound. Once a command cycle is completed, the upper bound of the force is updated as follows:
\begin{equation}
  \begin{aligned}
f^{\max}_{t+1} &= 
    \begin{cases}
        \min(f^{\max}_{t} + \Delta f, f^{\max}), & \text{if } \mathcal{C}, \\
        f^{\max}_{t}, & \text{otherwise},
    \end{cases}
  \end{aligned}
\end{equation}
where $f^{\max}$ denotes the maximum force goal for the curriculum training, and $\Delta f$ represents the incremental step.

\begin{figure}[!t]
\centering
\includegraphics[width=0.7\linewidth]{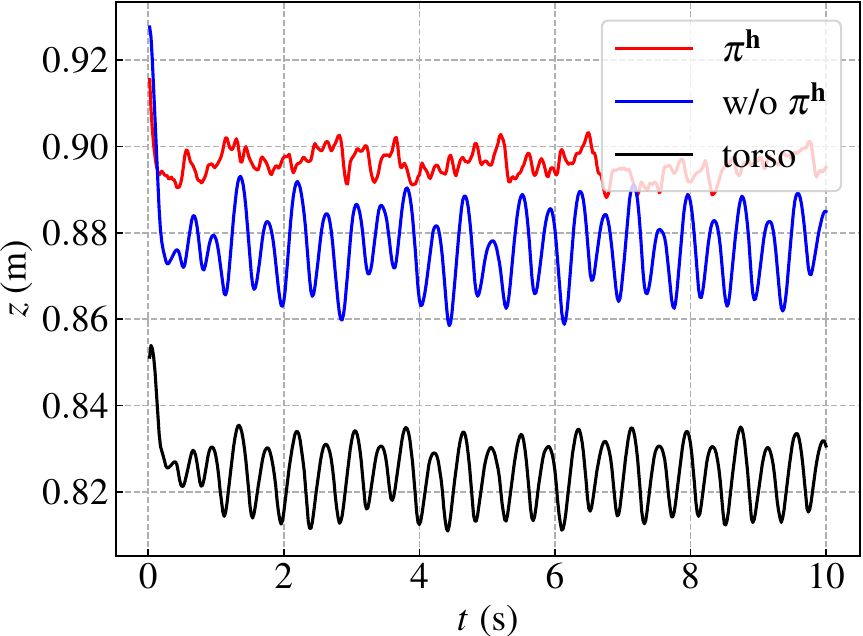}
\caption{Comparison of end-effector height trajectories in simulation, with (red) and without (blue) the delta-command policy. The black curve represents the torso height variation over time. During data collection, the robot is commanded to move along the $x$-axis at a velocity of \SI{0.7}{\meter\per\second}.}
\label{fig: fig_delta_wrist_torso}
\end{figure}

In the lower-body curriculum training, the condition $\mathcal{C}_{\text{dis}}$ ensures that the agent prioritizes acquiring basic locomotion skills before gradually learning to exert forces on the environment. The conditions $\mathcal{C}_{\text{lin}}$ and $\mathcal{C}_{\text{ang}}$ serve as evaluation metrics to ensure that the agent maintains adequate locomotion performance while progressing in force-capable training. Together, these three conditions enforce a staged training process for the lower-body policy. Moreover, the carefully designed force command sampling strategy, as defined in \eqref{eq:force_sample_strategy}, enhances the agent's generalization ability. The curriculum training progress exhibits a decelerating trend as the $f^{\max}_{t}$ is progressively increased. In the absence of the sampling strategy, the target force $f^{*}_t$
may remain consistently high values, causing the agent, after prolonged training under large-force targets, to converge toward a local optimum that only adapts to large forces (as illustrated in Fig.~\ref{fig: compare of force sample method}). Consequently, the proposed sampling strategy prevents the agent from forgetting previously acquired low-force capabilities, facilitating the acquisition of the force-capable skill.

% 最后，在下肢的训练过程中，上肢的policy也会保持运行，使得下肢策略能够泛化上肢运动带来的干扰
%在上述的力量课程训练方法中，条件Cdis 确保了agent会优先获取正常的locomotion技能，再开始学习对外界施加力的能力，条件C1和C2则能有效确保agent在力量训练的过程中，locomotion能力不会退化，保证了循序渐进的力量学习过程。同时，精心设计的力量cmd采样方法也保证了在训练过程中，若力量上限增长缓慢，agent不会收敛至只适应大力量的局部最优解。
% 为了达到force-capable的训练目标，下肢目标空间中增加了一个力量目标f，f的方向与机器人实际的vx保持一致或者相反，f的大小则通过一种设计的方在法[0,fmax]中采样。f确定后,根据牛顿第三定律，仿真环境中会产生一个与f方向相反的力作用于机器人的torso，来达到训练的效果。然而，在locomotion训练过程中持续施加外力是一种非常有挑战性的情况，如果采样的f远超策略控制能力能力，会导致策略难以收敛或者缓慢收敛，所以课程学习的方法被引入下肢训练过程，让策略可以顺序获得locomotion和force-capable的能力
% 力量课程训练的推进需要满足多项指标的评估，这些评估项被定义为：…… 力量指令f_t的采样方式如下：

\subsection{Whole Body Control via Delta-command Policy} 
\label{sec:delta-command}

The decoupled training of the upper- and lower-body policies provides the robot with effective loco-manipulation capabilities. However, the upper-body policy, which primarily learns end-effector pose tracking in the torso frame, is inevitably influenced by lower-body movements. Moreover, certain loco-manipulation tasks, such as cart-pushing, require the end-effector to maintain a constant height in the world frame. Therefore, the ability to regulate the end-effector pose directly in the world frame becomes indispensable. To address this limitation, a high-level delta-command policy is introduced. In contrast to the basic policy, this policy does not produce independent actions, but rather generates the upper-body policy's tracking objectives, thereby enabling the robot to acquire end-effector control ability in the world frame.

%delta-command agent的核心目标是学会控制机器人末端在世界坐标系中的高度和朝向保持恒定，不受下肢运动和其他外界因素的干扰，对于xy平面上的位姿则不做要求。因为对于loco-manipulation任务而言，若保持末端位置在xy平面不变等于几乎摒弃了机器人的移动能力。对此，delta-command policy 的定义如下

\begin{table}[!t]
\centering
\caption{Observations and Training Configurations}
\label{tab:obs and training cfg}
\resizebox{\linewidth}{!}{%
\begin{tabular}{lccc}
\toprule
\multicolumn{1}{c}{Term} & Upper-body & Lower-body & Delta-command \\ \midrule
Observation &
  $\bm{s}^{\text{u}}_{t}, \mathcal{G}^{\text{u}}_{t}$ &
  $\bm{s}^{\text{l}}_{t}, \mathring{{\bm{v}}}_{t}, \mathcal{G}^{\text{l}}_{t}$ &
  $\bm{s}_{t}, \mathring{{\bm{v}}}_{t}, \mathring{\mathbf{x}}^{\text{world}}_{t}, \mathcal{G}^{\text{h}}_{t}$ \\
Action scale             & 0.5        & 0.1        & 0.1           \\
Iterations       & 3000       & 15000      & 3000          \\ \midrule
Parallel envs.    & \multicolumn{3}{c}{4096}                \\
Actor hidden size       & \multicolumn{3}{c}{[512, 256, 128]}     \\
Critic hidden size      & \multicolumn{3}{c}{[512, 256, 128]}     \\
Activation function      & \multicolumn{3}{c}{ELU}                 \\ \bottomrule
\end{tabular}%
}
\end{table}

The core objective of the delta-command agent is to maintain the end-effector's height and roll-pitch orientation constant in the world frame, thereby mitigating disturbances caused by lower-body motions and other dynamic factors. In contrast, no constraints are imposed on the end-effector's displacement in the $xy$-plane or its yaw angle, since enforcing such a restriction in loco-manipulation tasks would almost eliminate the robot's locomotion capability. Based on this motivation, the delta-command policy is defined as follows:

% In the context of loco-manipulation tasks, preserving a fixed end-effector position in the world frame's $XY$-plane holds little practical significance. Consequently, leveraging the availability of ground-truth end-effector poses from the simulation environment, the core objective of the delta-command agent is to learn a control policy that keeps the end-effector's height and orientation constant in the world frame. This agent serves to counteract the disturbances induced by lower-body motion and other dynamic factors. The formal definition of the delta-command agent is given as follows:
\begin{equation}
\pi^{\text{h}}: \left \langle s_{t}, \bm{a}^{\text{h}}_{t-1} \right \rangle  \times \mathcal{G}^{\text{h}}_{t} \mapsto \bm{a}^{\text{h}}_{t} \in \mathbb{R}^{6}.
\end{equation}
Here, the action $\bm{a}^{\text{h}}_{t}$ represents the goal poses of both left and right end-effectors in terms of $z$-axis height and roll-pitch orientation, which are relative to world frame.
Accordingly, the upper-body goal is obtained by augmenting the baseline command $\mathcal{G}^{\text{u}}_{\text{ref}} \in \mathbb{R}^{12}$ with the delta-command $\bm{a}^{\text{h}}_{t} \in \mathbb{R}^{6}$:
% Accordingly, the upper-body command consists of the delta-command policy's action, along with independently specified $xy$-axis displacements and yaw rotation relative to the torso frame.
\begin{equation}
  \mathcal{G}^{\text{u}}_{t} = \mathcal{F}(\bm{a}^{\text{h}}_{t}, \mathcal{G}^{\text{u}}_{\text{ref}}),
\end{equation}
where $\mathcal{G}^{u}_{\text{ref}}$ denotes the baseline upper-body goal, and function $\mathcal{F}(\cdot)$ modifies the vector $\mathcal{G}^{\text{u}}_{\text{ref}}$. Specifically, it preserves all entries of $\mathcal{G}^{\text{u}}_{\text{ref}}$ with the exception of the six dimensions associated with the end-effectors' height and roll-pitch, where it applies an element-wise addition of the vector $\bm{a}^{\text{h}}_{t}$. Through this modification, the delta-command policy maintains the end-effectors' height and their roll-pitch orientations in the world frame, while adjustments of $\mathcal{G}^{u}_{\text{ref}}$ control the $xy$-axis displacements and yaw rotation relative to the torso frame.

Fig.~\ref{fig: fig_delta_wrist_torso} illustrates the effectiveness of the proposed delta-command policy. The three curves correspond to the torso height (black curve), the end-effector height without the policy (blue curve), and the end-effector height with the policy (red curve) in the world frame. The initial sharp drop is attributed to simulation initialization. During locomotion, the torso exhibits a periodic oscillatory pattern, which directly propagates to the end-effector. Without the delta-command policy, the end-effector follows this oscillation with even greater amplitude. In contrast, with the delta-command policy, the end-effector is regulated around the target height of \SI{0.9}{\meter}, effectively mitigating disturbances induced by lower-body motion. After excluding initialization transients, the oscillation amplitude of the end-effector is reduced from \SI{3.5}{\centi\meter} (without policy) to \SI{1.5}{\centi\meter} (with policy), corresponding to a \SI{57}{\percent} reduction.

\begin{figure}[!t]
\centering
\includegraphics[width=0.9\linewidth]{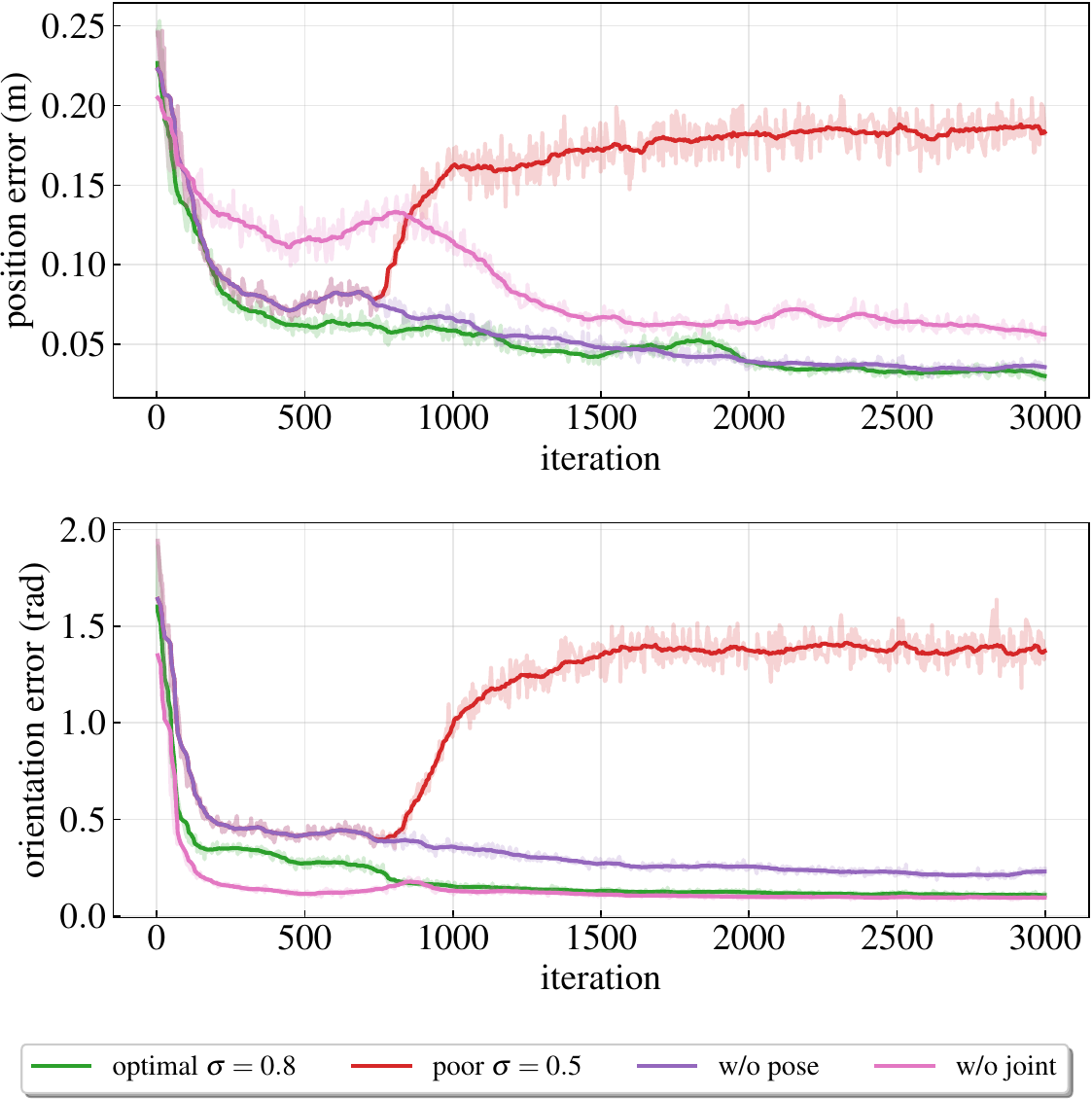}
\caption{Position and orientation errors of the left end-effector during training for four ablation groups: optimal $\sigma$ (heuristic reward with optimal $\sigma$), poor $\sigma$ (heuristic reward with undesirable $\sigma$), w/o pose (without task-space term), and w/o joint (without joint-space term).}
\label{fig: upper-body ablation}
\end{figure}

\section{Experiments and Results}\label{sec:experiment}
% 这一章节是对本文所提出的全身控制方法的实验验证，通过多种方式和场景证明了策略的有效性和泛化性。
%消融实验表明了上肢的启发式奖励函数和下肢的课程学习框架会给agent训练带来显著提升。最后，实体实验证明了我们的策略可以泛化sim2real之间的gap，并且actual environments中表现不错
This section presents the experimental validation of the proposed whole-body control framework. We evaluate the learned policies across diverse tasks and scenarios to demonstrate their effectiveness and generalization capability. Ablation studies further highlight that the heuristic reward design for upper-body control, and the curriculum learning strategy for lower-body training significantly accelerate and stabilize policy optimization. Finally, hardware experiments validate that our method successfully bridges the sim-to-real gap and achieves compelling performance in actual environments.

\subsection{Training}

\begin{table*}[!t]
\small
\centering
\caption{Reward Terms and Weights}
\label{tab:rewards}
\resizebox{\textwidth}{!}{%
\begin{tabular}{lcccc}
\toprule
\multicolumn{1}{c}{\multirow{2}{*}{Reward Term}} &
  \multirow{2}{*}{Function} &
  Upper-body &
  Lower-body &
  Delta-command \\ \cmidrule{3-5} 
\multicolumn{1}{c}{} &
   &
  \multicolumn{3}{c}{Weights} \\ \midrule
Linear velocity  tracking (root) & $\exp(-||\bm{v}^{*}_{xy}-\bm{v}_{xy}||^{2}_{2} / 0.5^{2})$ & - & 1 & - \\
Linear velocity penalty (root) & ${v}^{2}_{z}$ & - & -1 & - \\
Angular velocity tracking (root) & $\exp(-||{\omega}^{*}_{z}-{\omega}_{z}||^{2}_{2} / 0.5^{2} )$ & - & 1 & - \\
Angular velocity penalty (root) & $-||\bm{\omega}_{xy}||_{2}^{2}$ & - & -0.05 & - \\
Angle velocity penalty (head) & $-||\bm{\omega}^{\text{head}}_{xy}||_{2}^{2}$ & - & -0.5 & - \\
Termination penalty & $\mathds{1}[\text{termination}]$ & - & -200 & -200 \\
Joint acceleration penalty & $||\ddot{\bm{q}}||^{2}_{2}$ & - & -2.5e-7 & - \\
Joint torque penalty & $\sum_{i}| \bm{\tau}_{i} \cdot \dot{\bm{q}}_{i}|$ & - & -1e-3 & - \\
Joint position limits & $\sum_{i}\max(\bm{q}_{i} - 0.9 \cdot \bm{q}_{\text{lim}, i}, 0)$ & - & -2.0 & - \\
Collision penalty & $\mathds{1}[\text{collision}]$ & -1 & -1 & -1 \\
Straight body & $|| \bm{g}_{xy} ||^{2}_{2}$ & - & -1 & - \\
Feet air time~\cite{rudin2022learning} & $0.3 \cdot \min(T_{\text{air}}, 0.4)$ & - & 0.15 & - \\
Feet slide & $\sum_{i}\mathds{1}[\text{contact}_{i}] \cdot || \bm{v}^{\text{feet}}_{xy,i}||_{2}$ & - & -0.25 & - \\
Feet stumble & $\mathds{1}[|| \bm{F}^{\text{feet}}_{xy} || > 5 | {F}^{\text{feet}}_{z}|]$ & - & -2.0 & - \\
Feet force & $\min (\max ({F}^{\text{feet}}_{z} - 500, 0), 400)$ & - & -3e-3 & - \\
Joint deviation (waist) & $ \sum | \bm{q}^{\text{waist}}_{rp} - \bm{q}^{\text{default}} | $ & - & -1.2 & - \\
Joint deviation (hip) & $ \sum | \bm{q}^{\text{hip}}_{y} - \bm{q}^{\text{default}} | $ & - & -0.15 & - \\
Joint deviation (knee)    & $ \sum | \bm{q}^{\text{knee}}_{p} - \bm{q}^{\text{default}} | $ & - & -0.02 & - \\
Joint deviation (ankle)    & $ \sum | \bm{q}^{\text{ankle}}_{rp} - \bm{q}^{\text{default}} | $ & - & -0.02 & - \\
Action rate & $||\bm{a}_{t} - \bm{a}_{t-1}||^{2}_{2}$ & -1e-4 $\rightarrow$ -0.1 & -0.01 & -0.5 \\
Heuristic reward & (\ref{eq:heuristic reward}) & 1 & - & - \\
Joint velocity & $|| \dot{\bm{q}} ||^{2}_{2}$ & -1e-4 $\rightarrow$ -2e-3 & - & - \\
Action & $|| \bm{a}_{t} ||^{2}_{2}$ & - & - & -0.02 \\
End-effector height & $\left \| \mathbf{d}^{\text{world}*}_{z} - \mathbf{d}^{\text{world}}_{z} \right \|^{2}_{2}$ & - & - & -10 \\
End-effector orientation & $\sum_{i} || \text{Log}(\eta^{\text{world}}_{i} \otimes \hat{\eta}^{\text{\text{world}*}}_{i}) ||$ & - & - & -6 \\
\bottomrule
\end{tabular}%
}
\end{table*}

\begin{table*}[!t]
\small
\centering
\caption{Domain Randomization terms}
\label{tab:Domain Randomization}
\resizebox{\textwidth}{!}{%
\begin{tabular}{llclccc}
\toprule
\multicolumn{2}{c}{Term} &
  \multicolumn{2}{c}{Specification} &
  Upper-body &
  Lower-body &
  Delta-command \\ \midrule
\multicolumn{2}{l}{Static friction} &
  \multicolumn{2}{c}{$\mathcal{U}(0.6, 1.0)$} &
  $\surd$ &
  $\surd$ &
  $\surd$ \\
\multicolumn{2}{l}{Dynamic friction} &
  \multicolumn{2}{c}{$\mathcal{U}(0.4, 0.8)$} &
  $\surd$ &
  $\surd$ &
  $\surd$ \\
\multicolumn{2}{l}{Torso link mass} &
  \multicolumn{2}{c}{$\mathcal{U}(-5, 5) + \text{default}$~\si{\kilogram}} &
  - &
  $\surd$ &
  $\surd$ \\
\multicolumn{2}{l}{Hand link mass} &
  \multicolumn{2}{c}{$\mathcal{U}(1.0, 1.5) \times \text{default}$~\si{\kilogram}} &
  $\surd$ &
  $\surd$ &
  $\surd$ \\
\multicolumn{2}{l}{Root velocity} &
  \multicolumn{2}{c}{$\mathcal{U}(-0.5, 0.5)$~\si{\meter\per\second}} &
  - &
  $\surd$ &
  $\surd$ \\
\multicolumn{2}{l}{Joint position} &
  \multicolumn{2}{c}{$\mathcal{U}(0.5, 1.5) \times \bm{q}^{\text{default}}$~\si{\radian}} &
  $\surd$ &
  $\surd$ &
  $\surd$ \\ 
\multicolumn{2}{l}{Push robot} &
  \multicolumn{2}{c}{interval $\in [10, 15]$~\si{\second}, $\bm{v}_{xy} \in [-1, 1]$~\si{\meter\per\second}} &
  - &
  $\surd$ &
  $\surd$ \\
\multicolumn{2}{l}{Add force on hand} &
  \multicolumn{2}{c}{interval $\in [2, 5]$~\si{\second}, $F \in [0, 35] $~\si{\newton}} &
  $\surd$ &
  $\surd$ &
  $\surd$ \\ \bottomrule
\end{tabular}%
}
\end{table*}

\begin{table*}[!t]
\small
\centering
\caption{Upper-body Ablation Groups and Results}
\label{tab:upper-body ablation}
\resizebox{\textwidth}{!}{%
\begin{tabular}{@{}lccccccccc@{}}
\toprule
\multicolumn{1}{c}{\multirow{2}{*}{Training}} &
  \multicolumn{1}{c}{$\alpha^{\text{joint}}$} &
  \multicolumn{3}{c}{$\alpha^{\text{pose}}$} &
  \multicolumn{3}{c}{$\sigma$} &
  \multirow{2}{*}{Position error $\downarrow$} &
  \multirow{2}{*}{Orientation error $\downarrow$} \\ \cmidrule(lr){2-2} \cmidrule(lr){3-5} \cmidrule(lr){6-8}
\multicolumn{1}{c}{} & Stage I & Stage I & Stage II & Stage III & Stage I & Stage II & Stage III              &               \\ \midrule
Optimal $\sigma$     & 1       & 0    & -0.2   &-2    & 2       & 1        & 0.8       & \textbf{0.03} & \textbf{0.11} \\
Poor $\sigma$        & 1       & 0    & -0.2    &-2   & 2       & 0.5      & -         & 0.17          & 1.32          \\
w/o pose             & 1       & 0       & \multicolumn{2}{c}{-}    & 2       & 1        & 0.8       & 0.04          & 0.21          \\
w/o joint            & 0       & -0.2    & -2    &0   & \multicolumn{3}{c}{-}          & 0.05          & \textbf{0.11}           \\ \bottomrule
\end{tabular}%
}
\end{table*}

% 我们使用Isaacsim作为机器人训练和评估的仿真环境，训练跑在一个有着32核CPU和NVIDIA RTX4090 GPU 的高性能计算平台上。
% 由于实体机器人的状态信息是部分可观测的，我们采用了不对称actor-critic网络在下肢和delta-command agent的训练过程中，privileged information和训练的网络参数如表x所示。三个agent都采用纯正则化的奖励函数来充分释放策略的探索，如表x所示。最后，为了弥补sim和real之间的差距，表x所示域随机化项被施加在训练过程中。
We employ IsaacSim as the simulation environment for both training and evaluation of the humanoid robot. The training is conducted on a computing platform equipped with a AMD 7950X CPU and an NVIDIA RTX 4090D GPU.

The off-the-shelf RL algorithm, PPO~\cite{schulman2017proximal}, is adopted for policy optimization. Since the states of the physical robot are only partially observable in real-world, the asymmetric actor-critic network is adopted during the training of the lower-body and delta-command agents. The corresponding observations and the training configurations are summarized in Table~\ref{tab:obs and training cfg}, where $(\mathring{\ })$ indicates privileged information accessible only to the critics during training but not during deployment. 

Regarding the goal space, the upper-body targets are computed through forward kinematics following Alg.~\ref{alg:sampling joint pose}. The orientation aperture $\delta$ in task-space is set to $\pi$. 
For the lower body, the target linear velocity along the $x$-axis is sampled as 
${v}_x^{*} \sim \mathcal{U}(-0.6, 1.0)$~\si{\meter\per\second}, 
the target angular velocity around the $z$-axis as 
${\omega}^{*}_z \sim \mathcal{U}(-\frac{\pi}{2}, \frac{\pi}{2})$~\si{\radian}, 
and the maximum force goal $f^{\max}=50$~\si{\newton}. The target space of the delta-command is defined as 
$\mathbf{d}^{\text{world}*}_{z} \sim \mathcal{U}(0.85, 1)$~\si{\meter}, 
$\psi^{\text{world}*}_{r} \sim \mathcal{U}(-\frac{\pi}{2}, \frac{\pi}{2})$~\si{\radian}, and 
$\psi^{\text{world}*}_{p} \sim \mathcal{U}(-\frac{\pi}{2}, \frac{\pi}{2})$~\si{\radian}, 
which aims to maintain the end-effector's vertical displacement $\mathbf{d}^{\text{world}*}_{z}$ in world frame, together with the desired roll and pitch rotations, denoted as $\psi^{\text{world}*}_{r}$ and $\psi^{\text{world}*}_{p}$, respectively.
% 就目标空间而言，upper-body 的目标是根据alg.1中的采样方法后通过正运动学计算求得，G1任务空间的设定如图x所示。Lower-body的目标速度为vx~U(-0.6-1.0) ,vz=(-\frac{\pi}{2},\frac{\pi}{2}),ft=(0,50)。delta-command的目标空间是dz~U(0.85，100)，x~U(\frac{\pi}{2}, \frac{\pi}{2})y方向上的旋转能保持不变。

To encourage effective exploration, all three agents are trained with purely regularization-based reward functions, as detailed in Table~\ref{tab:rewards}. Here, the symbol ``${\mathds{1}}$'' denotes the indicator function, $\bm{q}^{i}_{rpy}$ are the roll-pitch-yaw rotational joints of body link $i$, $T_{\text{air}}$ is the single-foot air time at target speed exceeds \SI{0.1}{\meter\per\second}, $\bm{F}^{\text{feet}}_{j}$ represents force on feet along direction $j$, while ``$\rightarrow$'' indicates that the corresponding weight varies over the course of training.

Moreover, to bridge the gap between simulation and real-world, the domain randomization terms listed in Table~\ref{tab:Domain Randomization} are incorporated during the training process. To ensure that the robot can successfully exert force on the external environment, an additional force application term at the end-effectors of upper-body is introduced.

\begin{figure}[!t]
\centering
\includegraphics[width=0.7\linewidth]{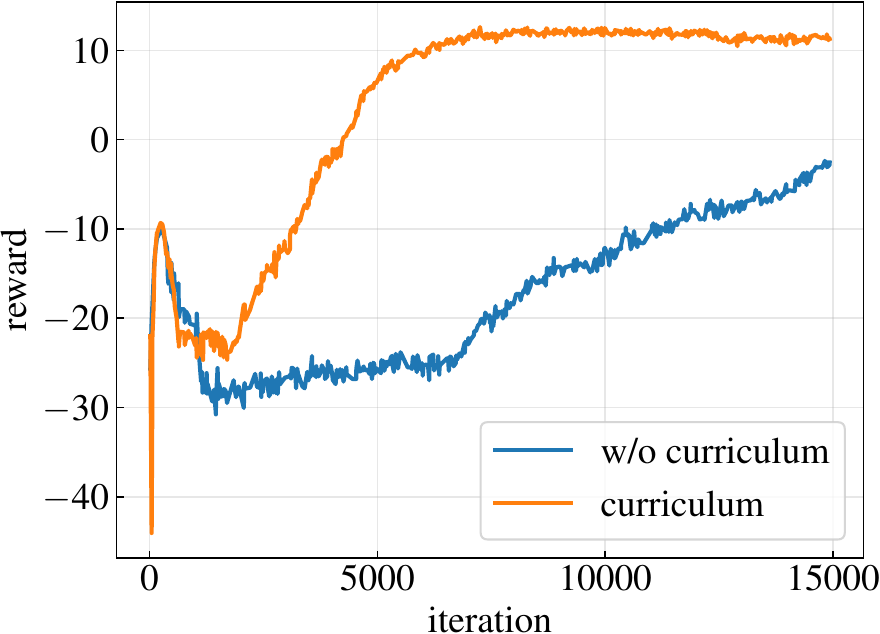}
\caption{Comparison of curriculum reward evolution between agents trained with and w/o curriculum learning.}
\label{fig: lower-body ablation}
\end{figure}

During the training of upper-body, the robot's torso link is fixed in the simulation environment, such that only the motions of the upper-body joints are considered. Besides, the training of the upper-body policy is conducted in stages. A sliding window of 100 episodes is employed, and if the reward values within the window vary by less than 1.5, the policy is considered to be approaching convergence, at which point the parameters $\alpha^{\text{pose}}$ and $\sigma$ are automatically adjusted and the training proceeds to the next stage. Reducing $\sigma$ or increasing $\alpha^{\text{joint}}$ both amplify the sensitivity of the reward to joint errors. However, owing to the exponential-form term, adjusting $\sigma$ provides a more efficient way to modulate this sensitivity. Consequently, $\alpha^{\text{joint}}$ is kept constant at 1, and only $\sigma$ is varied. 

% 下肢训练中，参数设置如下x=1,y=3,c=3.同时, the upper-body policy remains active, allowing the lower-body policy to generalize to disturbances induced by upper-body movements.
In lower-body training, the thresholds are set as $\varepsilon_{\text{lin}} = 0.3$, $\varepsilon_{\text{ang}} = 0.45$, $\varepsilon_{\text{dis}}=4$. Since domain randomization terms such as push robot are introduced, the tracking metrics for linear and angular velocity should not be designed overly strict. And the update step size $\Delta f$ is set to \SI{2}{\newton}. Meanwhile, the upper-body policy runs inference in the lower-body training process, enabling the policy to generalize to disturbances caused by the upper-body movements.

\subsection{Ablation Studies}

\begin{figure}[!t]
\centering
\subfloat[]{\includegraphics[width=0.4\linewidth]{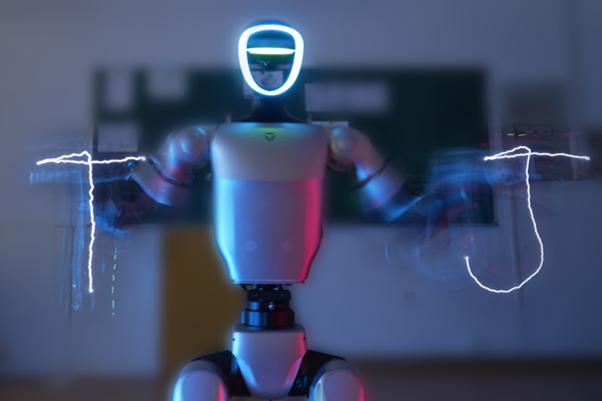} \label{fig: draw_TJ}}
\hfill
\subfloat[]{\includegraphics[width=0.52\linewidth]{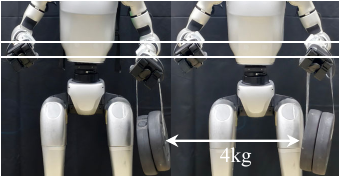} \label{fig: delta_command_force}}
\hfill
\subfloat[]{\includegraphics[width=0.29\linewidth]{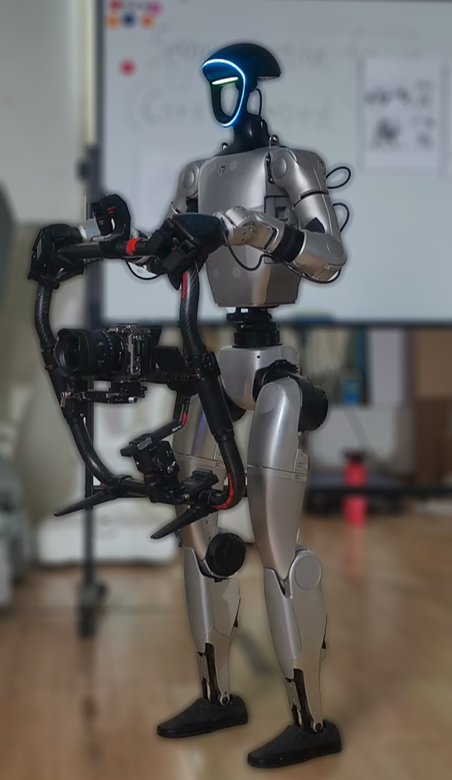} \label{fig: gimbal holding}}
\hfill
% \qquad \qquad \qquad
\subfloat[]{\includegraphics[width=0.64\linewidth]{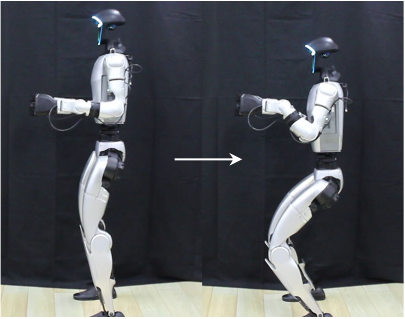} \label{fig: squat}}
\caption{Hardware experiment details. (a) The end-effector trajectories showed by LED light. (b) End-effector poses under a \SI{4}{\kilogram} payload. Without the delta-command policy (left), a clear downward displacement occurs, while with the policy (right), the end-effector is maintained closer to the desired pose. (c) The robot holds a gimbal. (d) The delta-command policy maintains the end-effector poses constant in the world frame while the robot squats.}
\label{fig:detail expressed hard-ware experiment}
\end{figure}

Ablation studies are conducted to systematically assess the contribution of each key component within the proposed training pipeline.

\textbf{Upper-body training pipeline.} Comparative experiments are conducted to evaluate the impact of the heuristic reward function. Based on different $\sigma$ update steps and module ablations, four ablation groups is defined, summarized in Table~\ref{tab:upper-body ablation}, where Stage I denotes the initial values.  Fig.~\ref{fig: upper-body ablation} visualizes the position and orientation errors of the left end-effector during training for all four groups. 

The results show that combining both joint- and task-space terms achieves the best performance. However, the “poor $\sigma$” group indicates that excessively large $\sigma$ update steps cause the heuristic reward to drop sharply, resulting in the agent focusing more on other reward terms, which in turn leads to an increase in both position and orientation errors. The “w/o pose” group shows that tracking only joint angles can still achieve reasonable position accuracy, but small angle errors in proximal joints propagate along the kinematic chain and degrade orientation tracking. In contrast, the “w/o joint” group reduces orientation error quickly, but without guidance from forward kinematics, the position error decreases slowly. Overall, these ablation results demonstrate the effectiveness of the heuristic reward, enabling faster policy convergence to higher upper-body performance.

\textbf{Lower-body training pipeline.}
%下肢训练中，我们针对课程训练进行消融实验，未使用课程训练的实验会在训练开始时就在进行力量采样：$f^{*}_{t}_\mathcal{U}$. 课程训练和非课程训练的奖励函数变化如图x所示。从奖励函数的上升趋势可以看出，课程训练的agent在xx步就呈现收敛趋势，而非课程训练的agent则是在最后都未呈现收敛趋势。因此，不使用课程训练时，agent的收敛速度明显变缓，这是因为在机器人还未习得基础的locomotion技能时就被赋予力量目标，使得仿真环境会施加与目标相反的力施加在机器人上，导致策略学习缓慢。此外，课程学习pipeline中采样方法的有效性已在图x时证明.、
Lower-body ablation is performed to evaluate the impact of curriculum learning. In the absence of curriculum training, the agent samples force targets at the beginning of training as $f^{*}_{t} \sim \mathcal{U}(0, f^{\max})$. In contrast, under curriculum training, force updates are restricted until \eqref{eq: force update condition} is satisfied.

The evolution of curriculum reward with and without curriculum learning is depicted in Fig.~\ref{fig: lower-body ablation}. The evolution indicate that, with curriculum learning, the policy converges as early as iteration 8000, whereas without curriculum learning, no convergence is observed within the entire training horizon. These findings suggest that curriculum learning substantially accelerates the convergence process during training. The underlying reason is that, without curriculum learning, the agent is required to track force targets before acquiring the fundamental locomotion skills, and the premature emphasis on force-capable skill severely impedes the learning of locomotion. By contrast, curriculum learning enables a progressive acquisition of skills from simple to complex, which proves to be highly effective. Furthermore, the efficacy of the sampling strategy within the curriculum learning framework has been previously validated, as illustrated in Fig.~\ref{fig: compare of force sample method}.

% 对于上肢训练框架，我们主要围绕启发式奖励函数展开对比实验。上肢训练过程中，一个长度为50的滑动窗口被设定，如果50个episodes中平均奖励在内变化低于1.5时则认为策略趋近收敛，此时会自动调整apose和std。std变小和a1增大都能增大joint-space 项的影响力，对此令a1恒等于1，只调节std。

%我们根据a1，a2，std的变化不同设置了4个对照组，如表x所示，Stage 表示对应参数发生了改变。由于左右末端的训练数据对称，我们将左侧末端的训练过程中四组实验的位置和朝向误差值可视化如图x所示。结合图和表可以看出，结合joint term和pose term的训练效果最佳，位置和朝向误差都比较小。但是poor sigma对照组表明joint term项对std变化比较敏感，若std变化幅度过大则会导致奖励值迅速下降趋近于0，使得agent被action rate 和joint velocity惩罚项吸引，导致追踪误差上升。从 w/o pose对照组可以看出，单纯的追踪joint angle也能达到不错的位置追踪效果，但是关节链中根部关节的微小误差会在关节末端处放大，导致朝向的追踪并不准确。此外，w/o joint对照组的朝向追踪误差快速下降，但是没有正运动学信息的指引，Position 误差下降缓慢。四个对照组的结果充分证明了heuristic reward的有效性，可以更快的让策略收敛至更好的效果。

\begin{figure}[!t]
\centering
\subfloat[]{\includegraphics[width=0.8\linewidth]{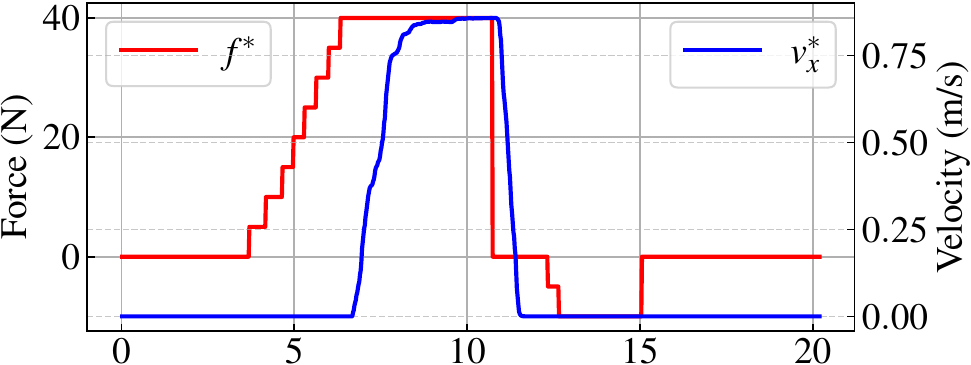} \label{fig: push_control_signal}}
\hfill
\subfloat[]{\includegraphics[width=0.84\linewidth]{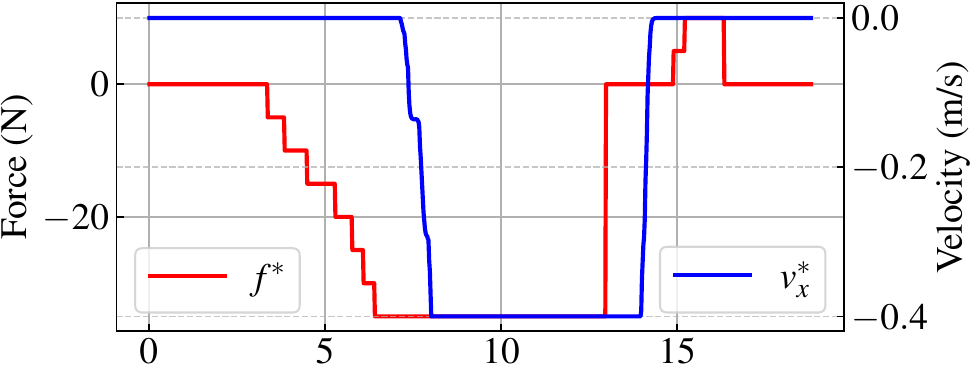} \label{fig:pull_control_signal}}
\caption{Time series of control signals in the cart-manipulation experiments. (a) Cart-pushing task. (b) Cart-pulling task.}
\label{fig: control signal}
\end{figure}

\subsection{Real-world Experiments}

%为了进一步验证提出方法的有效性，尤其是策略sim-to-real迁移能力和实际部署性能，我们开展了实体实验。

%在实体部署中，机器人的控制周期为50hz，三个policy可以同时在机器人板载的Jetson Orin NX上运行。policy的格式是executed in a Just-In-Time (JIT) mode with the C++ implementation of ONNX Runtime。
To further validate the effectiveness of the proposed approach, particularly its sim-to-real transferability and real-world deployment performance, hardware experiments are conducted on the Unitree G1 robot, which is equipped with two Dex3-1 dexterous hands at the upper-body end-effectors.
In the real-world deployment, the robot operates at a control frequency of \SI{50}{\hertz}, concurrently executing all three policies on the onboard Jetson Orin NX without additional fine-tuning. The policies are deployed in Just-In-Time (JIT) mode, with inference implemented in C++ via the ONNX Runtime to achieve real-time performance.

\begin{figure*}[!t]
\centering
\subfloat[]{\includegraphics[width=0.98\linewidth]{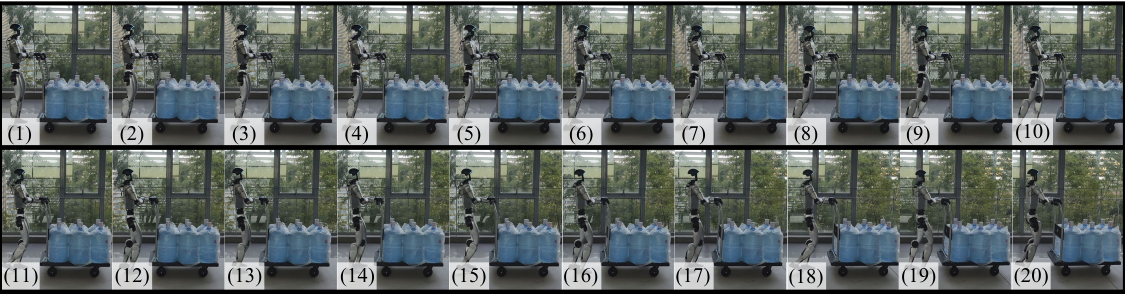} \label{fig: cart-pushing snapshot}}
\hfill
\subfloat[]{\includegraphics[width=0.98\linewidth]{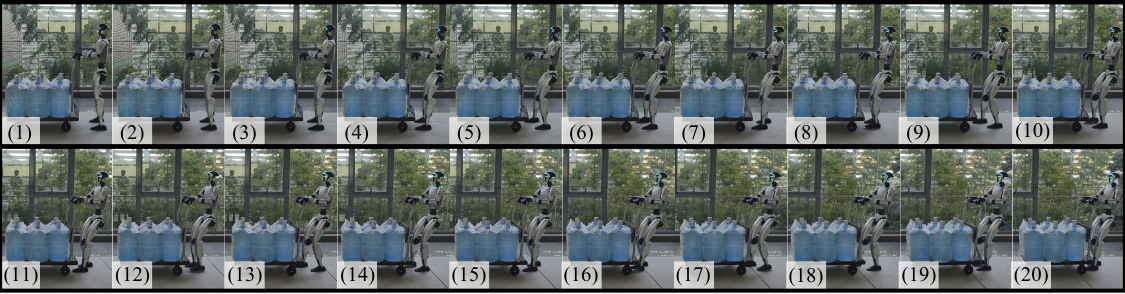} \label{fig: cart-pulling snapshot}}
\caption{Snapshots of the cart-manipulation experiments. (a) Cart-pushing task. (b) Cart-pulling task.}
\label{fig: cart-manipulation}
\end{figure*}

%图x展现了机器人上肢策略的运动能力，led灯珠被粘贴在手上如图x-（a）所示，通过延时摄影来展现出的末端轨迹。x-（b）中机器人左手追踪轨迹为字母“T”,右手追踪轨迹为字母“J”，每个字母的设计高度是20cm，宽度为10cm。从Led灯珠轨迹来看，机器人上肢策略的追踪效果不错。
A visualization of selected hardware experiment details is presented in Fig.~\ref{fig:detail expressed hard-ware experiment}. 
Here, Fig.~\ref{fig:detail expressed hard-ware experiment}\ref{sub@fig: draw_TJ} illustrates the upper-body policy's pose tracking capabilities. To visualize the end-effectors' trajectories, LED beads are affixed to both end-effectors, and their trajectory is recorded via long-exposure photography. Specifically, the robot's left hand is commanded to track a trajectory in the shape of the letter ``T'', while the right hand is commanded to track a trajectory in the shape of the letter ``J''. Both target letters are designed to be \SI{20}{\centi\meter} high and \SI{10}{\centi\meter} wide. The clear resulting trajectories from the LED beads provide qualitative evidence of the upper-body policy's tracking performance. Fig.~\ref{fig:detail expressed hard-ware experiment}\ref{sub@fig: delta_command_force} illustrates that the delta-command policy effectively mitigates the end-effector pose deviation caused by upper-body payloads. The robot end-effector is initialized to the same pose in both cases, with and without the delta-command policy, after which a \SI{4}{\kilogram} payload is attached to the end-effector. A clear difference in end-effector height is observed, which demonstrates that the delta-command policy enhances end-effector control in loco-manipulation tasks. Fig.~\ref{fig:detail expressed hard-ware experiment}\ref{sub@fig: gimbal holding} shows the robot holding a \SI{3.9}{\kilogram} camera gimbal while maintaining standing. This experiment demonstrates that the upper-body policy, after domain randomization, acquires the ability to accommodate payloads. Meanwhile, the lower-body policy generalizes to the additional upper-body load and preserves stable locomotion performance. 
Besides, Fig.~\ref{fig:detail expressed hard-ware experiment}\ref{sub@fig: squat} demonstrates the capability of the delta-command policy to maintain the end-effector pose in the world frame. The desired height of the end-effector is \SI{0.9}{\meter} in world frame, while its target roll and pitch orientation is zero. When the robot is commanded to squat by \SI{10}{\centi\meter}, the end-effector pose remains constant, effectively compensating for the downward displacement caused by knee flexion. 

Finally, to evaluate the collaborative performance of the three policies in loco-manipulation tasks, a cart-manipulation experiment reflecting an industrial scenario is conducted. 
The pushing and pulling time-varying control signals are depicted in Fig.~\ref{fig: control signal} and the snapshots of the robot action are shown in Fig.~\ref{fig: cart-manipulation}.
The experiment is set up on a moderately frictional plaster floor, with a payload of six water buckets (\SI{18.8}{\kilogram} each, totaling \SI{112.8}{\kilogram}) placed on the cart. The horizontal bar of the cart is \SI{0.9}{\meter} above the ground and serves as $\mathbf{d}^{\text{world}*}_{z}$ for the delta-command policy. Control signals are composed of continuous velocity commands, while the force commands are discretized and modified in \SI{5}{\newton} steps or reset to zero.
% $d_{z}$ in world frame, together with the desired roll and pitch rotations, denoted as $\psi^{\text{world}*}_{r}$ and $\psi^{\text{world}*}_{p}$, respectively

\textbf{Cart-pushing.} As shown in Fig.~\ref{fig: control signal}\ref{sub@fig: push_control_signal} and Fig.~\ref{fig: cart-manipulation}\ref{sub@fig: cart-pushing snapshot}, 
the robot remains stationary without receiving control commands during \num{0} to \SI{3.7}{\second}. From \num{3.7} to \SI{6.3}{\second}, the force command increases from \num{0} to \SI{40}{\newton}, during which the lower-body policy leverages its force-capable skill to actively apply force to the cart. This process corresponds to snapshots (1)--(7) in Fig.~\ref{fig: cart-manipulation}\ref{sub@fig: cart-pushing snapshot}, as the robot exhibits a noticeable forward lean without displacement. Once the cart begins to move, the velocity command is increased to \SI{0.8}{\meter\per\second}, prompting the robot to advance. During the stopping phase, the force and velocity commands are sequentially reduced to zero, after which the robot briefly applies a reverse force to counteract the cart's momentum.

\textbf{Cart-pulling.} The experiment details are shown in Fig.~\ref{fig: control signal}\ref{sub@fig:pull_control_signal} and Fig.~\ref{fig: cart-manipulation}\ref{sub@fig: cart-pulling snapshot}. 
During \num{3.4} to \SI{6.4}{\second}, the robot applies a pulling force of \SI{35}{\newton} through the force-capable skill. This process corresponds to snapshots (1)--(4) in Fig.~\ref{fig: cart-manipulation}\ref{sub@fig: cart-pulling snapshot}, where the robot exhibits a marked backward lean. Following the onset of cart motion, the velocity command varies to \SI{-0.4}{\meter\per\second}, and the robot engages in pulling. During stopping, the robot similarly applies a brief counterforce to assist the cart to rest.
% TODO
% Notably, the backward-leaning motion is preserved in the pulling experiment, whereas the forward-leaning motion is not retained in the pushing experiment. As a result, the robot applies a smaller active pulling force compared with pushing.
In addition, an interesting phenomenon in the experiments is observed. In the cart-pushing task, the forward-leaning posture that emerges when applying force (snapshot (7) in Fig.~\ref{fig: cart-manipulation}\ref{sub@fig: cart-pushing snapshot}) is not preserved during the subsequent movement (snapshots (12)--(20) in Fig.~\ref{fig: cart-manipulation}\ref{sub@fig: cart-pushing snapshot}). In contrast, in the cart-pulling task, the backward-leaning posture established during force exertion (snapshot (4) in Fig.~\ref{fig: cart-manipulation}\ref{sub@fig: cart-pulling snapshot}) is maintained throughout the following motion (snapshots (5)--(20) in Fig.~\ref{fig: cart-manipulation}\ref{sub@fig: cart-pulling snapshot}). This backward inclination enables the robot to pull the cart with a smaller active force compared to pushing.

Overall, the coordinated operation of the three policies enables the robot to perform high-payload cart-manipulation tasks, demonstrating the effectiveness of the proposed approach in loco-manipulation and its potential for industrial applications.

% Finally, 为了测试三个agent一起配合执行loco-manipulation任务的极限性能。一个贴近工业化场景运用的cart-manipulation实验被执行。机器人推/拉车的snapshot如图x所示，推/拉车实验中的target信号输入如图y所示。实验场景设计如下：机器人在摩擦力适中的石膏地面上manipulates车，车上负载6桶水，每桶水18.8kg，共112.8kg。cart的横杆高度为0.9m，作为目标输入给delta-command policy。控制信号的输入方式为线速度目标连续输入，力量目标按5N的步长上升/下降，或者归0。

%推车实验如图x和图y所示，机器人在0-3.7s内未收到控制指令保持静止，在3.7~6.3s力量目标从0增长至40N，这是运用下肢agent force-capable的技能主动对小车施加力。该过程对应图x中的（1）~（7），可以看出机器人在主动施加力时身体明显前倾但并未移动。小车被推动后，速度指令才增长至0.8m/s，机器人开始前进.停止时力量指令和速度指令先后归0，随后机器人短暂施加反向力来抵消小车惯性。

%拉车实验对应图x和y，3.4~6.4s中，机器人通过force-capable技能主动对小车施加35N的拉力。该过程对应图x中（1）~（5）,机器人施加拉力时有明显后倾趋势。当小车开始移动后，速度力量才降低至-0.4m/s,机器人开始拉车.停止时，机器人同样通过短暂施加反方向的力帮助小车停止。值得一提的是，拉车过程中，机器人的后倾动作得到了保留，而推车过程中前倾则没有得到保留，所以机器人在拉车时主动施加的力更小。综合来看，三个策略相互配合可以赋予机器人完成高负载的cart-manipulation任务，是本文提出方法在loco-manipulation的有力表现，也证明该方法在工业领域具有强大潜力。

\section{Conclusion}\label{sec:conclusion}
% 本研究提出了一种新颖的自适应轨迹优化方法，该方法将传统的基于优化的运动规划算法与深度强化学习相结合，以增强自主移动机器人的环境适应性。深度强化学习策略会根据环境背景动态调整优化目标，然后轨迹优化会利用这些目标生成高效安全的轨迹。在模拟和实际环境中进行的大量实验表明，所提出的算法在效率和防碰撞方面优于现有的DWA、TEB和端到端方法等方法。将经典的运动规划框架与基于学习的方法相结合已被证明能够有效地缩小模拟与现实之间的差距，而这种差距是将基于学习的机器人系统部署到实际场景中的关键障碍。除了基于优化的方法之外，我们未来的工作将侧重于进一步将经典方法（例如基于采样和基于搜索的策略）集成到基于学习的导航框架中。此外，我们的研究中也观察到了一些失败的案例。由于当前系统使用单线二维激光雷达，某些明显高于或低于扫描平面的障碍物无法被检测到。这种限制可能会增加此类场景下的碰撞风险。在未来的工作中，我们的目标是融入三维激光雷达传感技术，以实现更全面的环境感知。

%本文基于强化学习针对人形机器人loco-manipulation问题提出了一种多策略解决框架。该框架共包含upper-body ，  lower-body 和delta-command 三个agent。我们在upper-body agent 的训练中引入了启发式奖励函数，使策略能在正运动学信息的引导下快速收敛并达到更好的末端追踪效果。lower-body agent 则具有对环境主动施加力的特点，我们为其精心设计了一套课程学习训练方案，该方案能保证agent的技能学习顺序并大幅度提升策略收敛效率。此外，delta command agent 不直接参与机器人的控制，而是通过修改上肢agent的追踪目标来达到末端追踪世界坐标系位姿的能力。仿真环境和现实世界的大量实验结果真实了本文方法的有效性。尤其是在推车实现中，三个策略的有机融合使机器人可以推动和拉动112.8kg负重的小车，充分证明了该框架在工业场景的应用前景。

% 但是在目前的框架下，机器人只能向前或向后对外界施加力，因此未来的工作将会继续围绕force-capable展开, 增加机器人对力的控制能力。
This paper presents a multi-policy framework based on RL to tackle humanoid loco-manipulation tasks in high-load industrial scenarios. The proposed framework consists of three distinct policies: an upper-body policy, a lower-body policy, and a delta-command policy. For the upper-body policy, a heuristic reward function is introduced to guide the policy with forward kinematics priors. This reward significantly accelerates policy convergence and improves end-effector tracking performance. The lower-body policy is endowed with the capability of actively exerting forces on the environment. A curriculum learning scheme is meticulously designed for this policy, which ensures a sequential acquisition of skills and substantially improves training efficiency. Meanwhile, the delta-command policy, rather than directly controlling the robot, generates the tracking targets of the upper-body policy to achieve end-effector pose tracking in world-frame.

We conduct extensive evaluations in both simulation and real-world settings to examine the proposed approach. Notably, in the cart-manipulation task, the coordinated integration of the three policies enabled the G1 humanoid robot to push and pull a cart with a payload of \SI{112.8}{\kilogram}, demonstrating the framework's strong potential for industrial applications.

Future work will focus on expanding the robot's force control capabilities, as the current framework is limited to applying forces in a forward or backward direction.

\bibliographystyle{IEEEtran}
\bibliography{references}  

\newpage


% \end{IEEEbiography}



\vfill

\end{document}